\documentclass[10pt]{article}

\usepackage[left=1.25in,top=1.25in,right=1.25in,bottom=1.25in,head=1.25in]{geometry}
\usepackage{amsfonts,amsmath,amssymb,amsthm}
\usepackage{verbatim,float,url}
\usepackage{graphicx,subcaption,psfrag}
\usepackage{dsfont,bm}
\usepackage{natbib}

\usepackage{setspace}

\theoremstyle{remark}

\theoremstyle{definition}

\newcommand{\argmin}{\mathop{\mathrm{argmin}}}

\newcommand{\st}{\mathop{\mathrm{subject\,\,to}}}

\def\half{\frac{1}{2}}

\def\diag{\mathrm{diag}}

\def\hbeta{\hat{\beta}}

\def\htheta{\hat{\theta}}

\def\hu{\hat{u}}
\def\hf{\hat{f}}

\def\R{\mathbb{R}}

\def\DP{\mathrm{DP}}

\title{Fast and Flexible ADMM Algorithms for Trend Filtering} 

\author{
Aaditya Ramdas$^{21}$\\
\texttt{aramdas@cs.cmu.edu}
\and
Ryan J. Tibshirani$^{12}$\\
\texttt{ryantibs@stat.cmu.edu}\\
\and
$^1$Department of Statistics and $^2$Machine Learning Department\\
Carnegie Mellon University\\
}

\date{}

\begin{document}
\maketitle


\begin{abstract}
This paper presents a fast and robust algorithm for trend
filtering, a recently developed nonparametric regression tool.  It has
been shown that, for estimating functions whose derivatives are of
bounded variation, trend filtering achieves the minimax optimal error
rate, while other popular methods like smoothing splines and kernels
do not.  Standing in the way of a more widespread practical adoption,
however, is a lack of scalable and numerically stable algorithms for
fitting trend filtering estimates.  This paper presents a highly efficient,
specialized ADMM routine for trend filtering.  Our
algorithm is competitive with the specialized interior point methods
that are currently in use, and yet is far more numerically robust.
Furthermore, the proposed ADMM implementation is
very simple, and importantly, it is flexible enough to extend to
many interesting related problems, such as sparse trend filtering
and isotonic trend filtering.  Software for our method is freely
available, in both the C and R languages.

\end{abstract}
 

\section{Introduction}
\label{sec:intro}

Trend filtering is a relatively new method for nonparametric 
regression, proposed independently by \citet{hightv,l1tf}.  
Suppose that we are given output points $y=(y_1,\ldots y_n) 
\in \R^n$, observed across evenly spaced input points $x=(x_1,\ldots 
x_n) \in \R^n$, say, $x_1=1, \ldots x_n=n$ for simplicity. 
The trend filtering estimate  
\smash{$\hbeta=(\hbeta_1,\ldots \hbeta_n) \in \R^n$}
of a specified order $k\geq 0$ is defined as
\begin{equation}
\label{eq:tf}
\hbeta = \argmin_{\beta\in\R^n} \, \half \|y-\beta\|_2^2 +  
\lambda \|D^{(k+1)} \beta\|_1.
\end{equation}
Here $\lambda \geq 0$ is a tuning parameter, and $D^{(k+1)} \in
\R^{(n-k)\times n}$ is the discrete difference (or derivative)
operator of order $k+1$.  We can define these operators recursively as    
\begin{equation}
\label{eq:d1}
D^{(1)} = \left[\begin{array}{rrrrrr}
-1 & 1 & 0 & \ldots & 0 & 0 \\
0 & -1 & 1 & \ldots & 0 & 0 \\
\vdots & & & & & \\
0 & 0 & 0 & \ldots & -1 & 1
\end{array}\right],
\end{equation}
and 
\begin{equation}
\label{eq:dk}
D^{(k+1)} = D^{(1)} D^{(k)} \;\;\; \text{for}\;\, k=1,2,3,\ldots.
\end{equation}
(Note that, above, we write $D^{(1)}$ to mean the
$(n-k-1)\times (n-k)$ version of the 1st order difference matrix in
\eqref{eq:d1}.)
When $k=0$, we can see from the definition of $D^{(1)}$ in
\eqref{eq:d1} that the trend filtering problem \eqref{eq:tf} is the
same as the 1-dimensional fused lasso problem \citep{fuse}, also
called 1-dimensional total variation denoising \citep{tv}, and hence
the 0th order trend filtering estimate \smash{$\hbeta$} is 
piecewise constant across the input points $x_1,\ldots x_n$. 

For a general $k$, the $k$th order trend filtering estimate
has the structure of a $k$th order
piecewise polynomial function, evaluated across the inputs
$x_1,\ldots x_n$. 
The knots in this piecewise polynomial are selected adaptively 
among $x_1,\ldots x_n$, with a higher value of the tuning parameter
$\lambda$ (generally) corresponding to fewer knots.  
To see examples, the reader can jump ahead to the next subsection, or
to future sections.  For arbitrary input 
points $x_1,\ldots x_n$ (i.e., these need not be evenly spaced), the
defined difference operators will have different nonzero entries, but
their structure and the recursive relationship between them is
basically the same; see Section \ref{sec:uneven}.   

Broadly speaking, nonparametric regression is a well-studied field
with many celebrated tools, and so one may wonder about the merits of
trend filtering in particular.   For detailed motivation, we refer the
reader to \citet{trendfilter}, where it is argued that trend filtering 
essentially balances the strengths of smoothing splines
\citep{deboorsplines,wahbasplines} and locally
adaptive regression splines \citep{locadapt}, which are two of the
most common tools for piecewise polynomial estimation.  In short:   
smoothing splines are highly computationally efficient but are not
minimax optimal (for estimating functions whose derivatives are of
bounded variation); locally adaptive regression splines are minimax
optimal but are relatively inefficient in terms of computation;
trend filtering is both minimax optimal and computationally comparable
to smoothing splines.  \citet{trendfilter} focuses mainly on the
statistical properties trend filtering estimates,
and relies on externally derived algorithms for comparisons of
computational efficiency.

\subsection{Overview of contributions}

In this paper, we propose a new algorithm for trend
filtering. We do not explicitly address the problem of model
selection, i.e., we do not discuss how to choose the tuning parameter
$\lambda$ in \eqref{eq:tf}, which is a long-standing statistical 
issue with any regularized estimation method.  Our concern is
computational; if a practitioner wants to solve the trend filtering
problem \eqref{eq:tf} at a given value of $\lambda$ (or sequence of
values), then we provide a scalable and efficient means of doing so.
Of course, a fast algorithm such as the one we provide can still be 
helpful for model selection, in that it can provide speedups for
common techniques like cross-validation.     

For 0th order trend filtering, i.e., the 1d fused lasso
problem, two direct, linear time algorithms already exist: the first
uses a taut string principle \citep{tautstring}, and the second
uses an entirely different dynamic programming approach
\citep{nickdp}.  Both are extremely (and equally) fast in practice,
and for this special $0$th order problem, these two direct algorithms
rise above all else in terms of computational efficiency and numerical
accuracy.
As far as we know (and despite our best attempts), these algorithms
cannot be directly extended to the higher order cases
$k=1,2,3,\ldots$.  However, our proposal {\it indirectly} extends
these formidable algorithms to the higher order cases with a
special implementation of the alternating  direction method of
multipliers (ADMM). In general, there can be multiple ways to
reparametrize an unconstrained optimization problem 
so that ADMM can be applied; for the trend filtering problem
\eqref{eq:tf}, we choose a particular parametrization suggested by the
recursive decomposition  \eqref{eq:dk}, leveraging the fast, exact
algorithms that exist for the $k=0$ case. We find that this choice
makes a big difference in terms of the convergence of the resulting
ADMM routine, compared to what may be considered the standard ADMM
parametrization for \eqref{eq:tf}.

Currently, the specialized primal-dual interior point (PDIP) method 
of \citet{l1tf} seems to be the preferred method for computing trend 
filtering estimates. The iterations of this algorithm are
cheap because they reduce to solving banded linear systems (the
discrete difference operators are themselves banded).  
Our specialized ADMM implementation and the PDIP method have distinct
strengths. We summarize our main findings below.   
\begin{itemize}
\item Our specialized ADMM implementation converges more reliably
  than the PDIP method, over a wide range of
  problems sizes $n$ and tuning parameter values $\lambda$. 
\item In particular setups---namely, small problem sizes, and small
  values of $\lambda$ for moderate and large problem
  sizes---the PDIP method converges to high
  accuracy solutions very rapidly.  In such situations, our
  specialized ADMM algorithm does not match the convergence rate of
  this second-order method. 
\item However, when plotting the function estimates, our
  specialized ADMM implementation produces solutions of visually
  perfectly acceptable accuracy after a small number of
  iterations.  This is true over a broad range of problem sizes $n$ and 
  parameter values $\lambda$, and covers the settings in which its
  achieved criterion value has not converged at the rate of the
  PDIP method.    
\item Furthermore, our specialized ADMM implementation displays a
  greatly improved convergence rate over what may be thought of as the
  ``standard'' ADMM implementation for problem \eqref{eq:tf}.  Loosely
  speaking, standard implementations of ADMM are generally considered 
  to behave like first-order methods \citep{admm}, whereas our
  specialized implementation exhibits performance somewhere in between
  that of a first- and second-order method.
\item One iteration of our specialized ADMM implementation has linear
  complexity in the problem size $n$; this is also true for
   PDIP. Empirically, an iteration of our ADMM
  routine runs about 10 times faster than a PDIP iteration.
\item Our specialized ADMM implementation is quite simple
  (considerably simpler than the specialized primal-dual interior
  point method), and is flexible enough that it can be extended to
  cover many variants and extensions of the basic trend filtering
  problem \eqref{eq:tf}, such as sparse trend filtering, mixed trend 
  filtering, and isotonic trend filtering.
\item Finally, it is worth remarking that extensions beyond the 
  univariate case are readily available as well, as univariate
  nonparametric regression tools can be used as building blocks for
  estimation in broader model classes, e.g., in generalized additive
  models \citep{gam}.  
\end{itemize}

Readers well-versed in optimization may wonder about alternative 
iterative (descent) methods for solving the trend filtering problem
\eqref{eq:tf}. Two natural candidates that have enjoyed much success
in lasso regression problems are proximal gradient and coordinate
descent algorithms. Next, we give a motivating case study that 
illustrates the inferior performance of both of these methods for
trend filtering.  In short, their performance
is heavily affected by poor conditioning of the difference
operator $D^{(k+1)}$, and their
convergence is many orders of magnitude worse than the specialized
primal-dual interior point and ADMM approaches.

\subsection{A motivating example}
\label{sec:motivate}

Conditioning is a subtle but ever-present issue faced by iterative
(indirect) optimization methods.  This issue affects some algorithms 
more than  others; e.g., in a classical optimization 
context, it is well-understood that the convergence bounds for
gradient descent depend on the smallest and largest eigenvalues of the
Hessian of the criterion function, while those for Newton's method do 
not (Newton's method being affine invariant).  Unfortunately,
conditioning is a very real issue when solving the trend filtering
problem in \eqref{eq:tf}---the discrete derivative operators
$D^{(k+1)}$, $k=0,1,2,\ldots$ are extremely ill-conditioned, and this
only worsens as $k$ increases.  

\begin{figure}[htbp]
\centering
\includegraphics[width=0.48\textwidth]{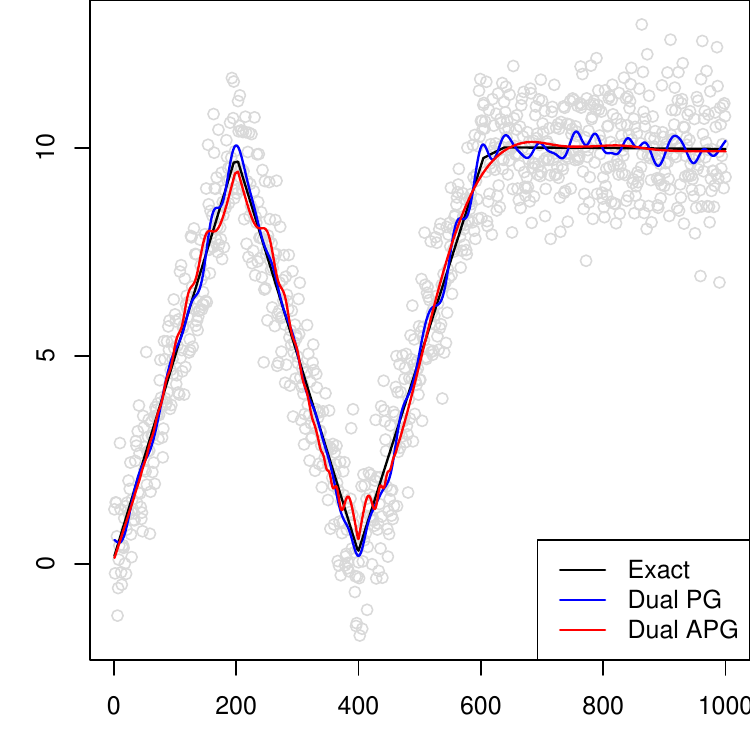}
\includegraphics[width=0.48\textwidth]{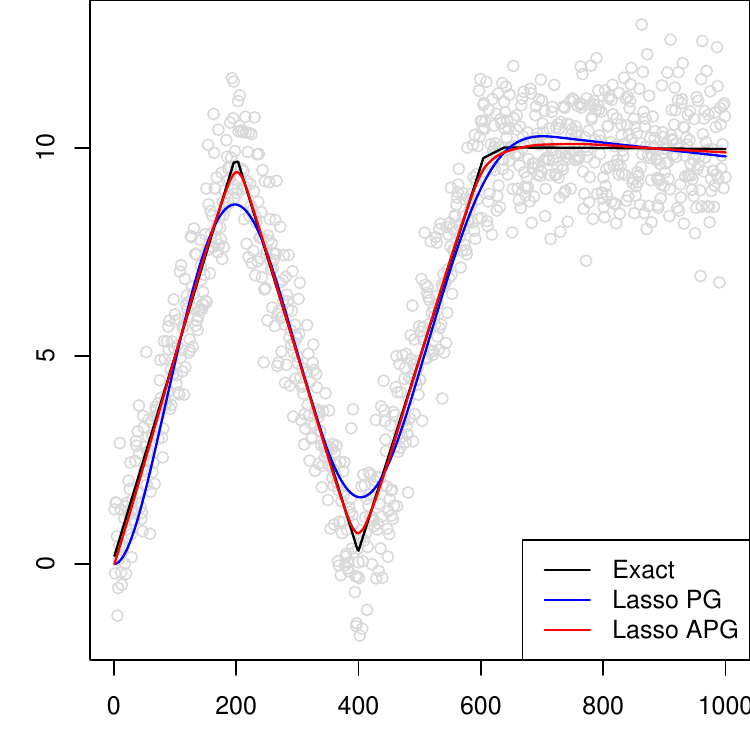} 
\includegraphics[width=0.48\textwidth]{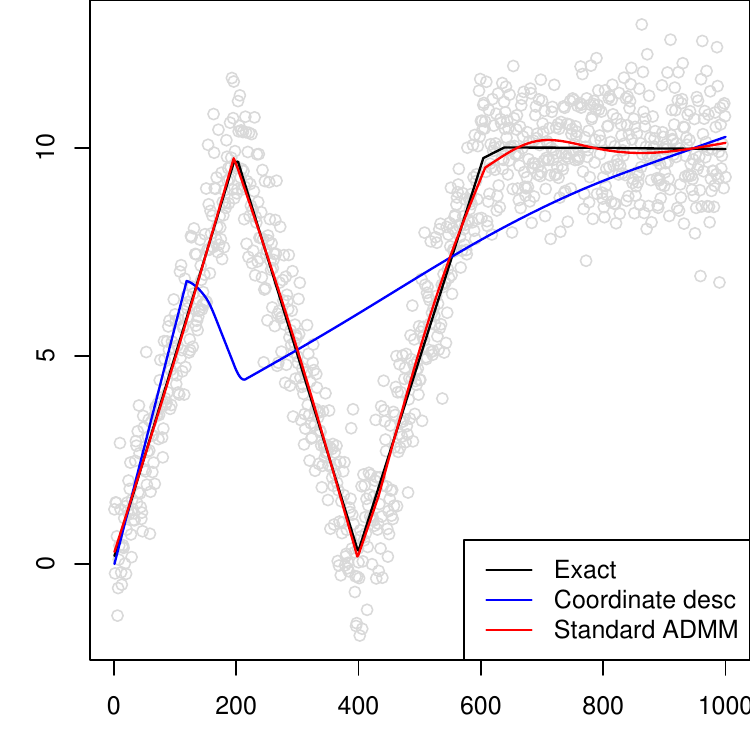}
\includegraphics[width=0.48\textwidth]{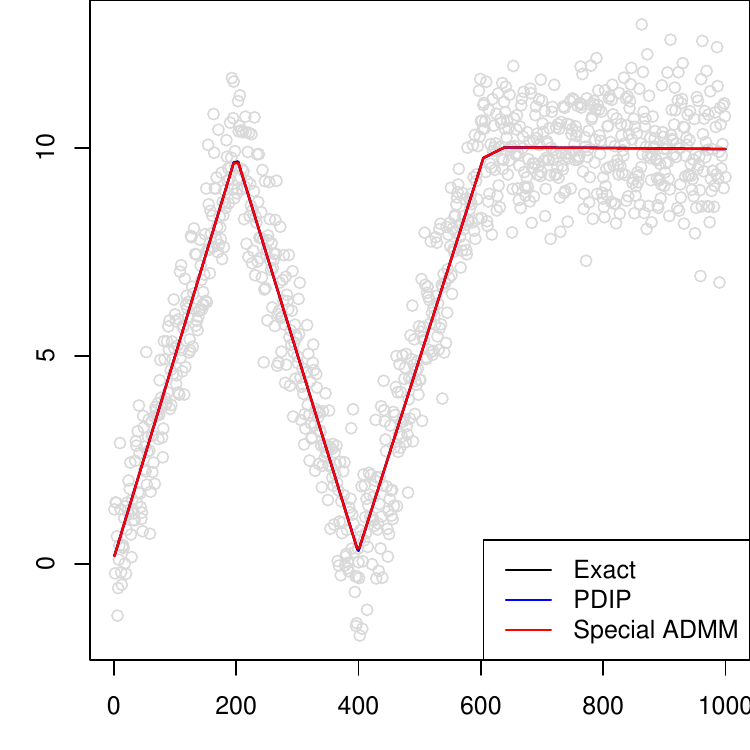}
\caption{\small\it
  All plots show $n=1000$ simulated observations in
  gray and the exact trend filtering solution as a black line, computed
  using the dual path algorithm of \citet{genlasso}.  The top left
  panel shows proximal gradient descent and its accelerated version
  applied to the dual problem, after 10,000 iterations.  The top
  right show proximal gradient and its accelerated version after
  rewriting trend filtering in lasso form, again after 10,000
  iterations.  The bottom left shows coordinate descent applied to the
  lasso form, and a standard ADMM approach applied to the original
  problem, each using 5000 iterations (where one iteration for
  coordinate descent is one full cycle of coordinate updates).   
  The bottom right panel shows the specialized PDIP and ADMM
  algorithms, which only need 20 iterations, and match the exact
  solution to perfect visual accuracy.  Due to the special form of the
  problem, all algorithms here have $O(n)$ complexity per iteration
  (except coordinate descent, which has a higher iteration cost).} 
\label{fig:motivate}
\end{figure}

This worry can be easily realized in examples, as we now demonstrate  
in a simple simulation with a reasonable polynomial order, $k=1$, and  
a modest problem size, $n=1000$.  For solving the trend filtering
problem \eqref{eq:tf}, with $\lambda=1000$, we compare proximal
gradient descent and accelerated proximal gradient method (performed 
on both the primal and the dual problems), coordinate descent, a
standard ADMM  approach, our specialized ADMM approach, and the
specialized PDIP method of \citet{l1tf}.  Details of the simulation
setup and these various algorithms are given in Appendix
\ref{app:motivate}, 
but the main message can be seen from Figure \ref{fig:motivate}. 
Different variants of proximal gradient methods, as well as
coordinate descent, and a standard ADMM approach, all perform quite
poorly in computing trend filtering estimate, but the second-order
PDIP method and our specialized ADMM implementation perform
drastically better---these two reach in 20 iterations what the others
could not reach in many thousands.  Although the latter two
techniques perform similarly in  this example, we will  
see later that our specialized ADMM approach generally suffers from
far less conditioning and convergence issues than PDIP, especially in
regimes of regularization (i.e., ranges of $\lambda$ values) that are
most interesting statistically.

The rest of this paper is organized as follows.  In Section
\ref{sec:admm}, we describe our specialized ADMM
implementation for trend filtering.  In Section \ref{sec:compare}, we
make extensive comparisons to PDIP.  
Section \ref{sec:uneven} covers the case of general input points
$x_1,\ldots x_n$. Section \ref{sec:extend} considers several
extensions of the basic trend filtering model, and the accompanying
adaptions of our specialized ADMM algorithm.  
Section \ref{sec:discuss} concludes with a short discussion.

\section{A specialized ADMM algorithm} 
\label{sec:admm}

We describe a specialized ADMM algorithm for trend
filtering.  This algorithm may appear to only slightly differ in its
construction from a more standard ADMM algorithm for trend
filtering, and both approaches have virtually the same computational 
complexity, requiring $O(n)$ operations per iteration; however, as we
have glimpsed in Figure \ref{fig:motivate}, the difference in
convergence between the two is drastic.     
 
The standard ADMM approach (e.g., \citet{admm}) is based on rewriting 
problem \eqref{eq:tf} as  
\begin{equation}
\label{eq:tf1}
\min_{\beta \in \R^n, \, \alpha \in \R^{n-k-1}} \,
\half \|y-\beta\|_2^2 + \lambda \|\alpha\|_1 \;\;\st\;\; 
\alpha = D^{(k+1)}\beta. 
\end{equation}
The augmented Lagrangian can then be written as
\begin{equation*}
L(\beta,\alpha,u) = \half \|y-\beta\|_2^2 + \lambda \|\alpha\|_1 +
\frac{\rho}{2} \|\alpha - D^{(k+1)}\beta + u \|_2^2 -
\frac{\rho}{2}\|u\|_2^2,
\end{equation*}
from which we can derive the standard ADMM updates:
\begin{align}
\label{eq:beta1} 
\beta &\leftarrow 
\big(I + \rho (D^{(k+1)})^T D^{(k+1)}\big)^{-1}
\big(y + \rho (D^{(k+1)})^T (\alpha + u)\big), \\ 
\label{eq:soft}
\alpha &\leftarrow 
S_{\lambda/\rho} (D^{(k+1)}\beta - u), \\ 
\label{eq:u1}
u &\leftarrow u + \alpha - D^{(k+1)}\beta.
\end{align}
The $\beta$-update is a banded linear system solve, with bandwidth
$k+2$, and can be implemented in time $O(n(k+2)^2)$ (actually,
$O(n(k+2)^2)$ for the first solve, with a banded Cholesky, and
$O(n(k+2))$ for each subsequent solve).   The
$\alpha$-update, where $S_{\lambda/\rho}$ denotes coordinate-wise
soft-thresholding at the level $\lambda/\rho$, takes time
$O(n-k-1)$.  The dual update uses a banded matrix multiplication,
taking time $O(n(k+2))$, and therefore one full iteration of
standard ADMM updates can be done in linear time (considering $k$ 
as a constant).  

Our specialized ADMM approach instead begins by rewriting
\eqref{eq:tf} as 
\begin{equation}
\label{eq:tf2}
\min_{\beta \in \R^n, \, \alpha \in \R^{n-k}} \,
\half \|y-\beta\|_2^2 + \lambda \|D^{(1)}\alpha\|_1 \;\;\st\;\; 
\alpha = D^{(k)}\beta,
\end{equation}
where we have used the recursive property $D^{(k+1)}=D^{(1)}D^{(k)}$.
The augmented Lagrangian is now
\begin{equation*}
L(\beta,\alpha,u) = \half \|y-\beta\|_2^2 + \lambda
\|D^{(1)}\alpha\|_1 + \frac{\rho}{2} 
\|\alpha - D^{(k)}\beta + u \|_2^2 - \frac{\rho}{2}\|u\|_2^2,
\end{equation*}
yielding the specialized ADMM updates:
\begin{align} 
\label{eq:beta2}
\beta &\leftarrow 
\big(I + \rho (D^{(k)})^T D^{(k)}\big)^{-1}
\big(y + \rho (D^{(k)})^T (\alpha + u)\big), \\ 
\label{eq:alpha2}
\alpha &\leftarrow 
\argmin_{\alpha \in \R^{n-k}} \,
\half \|D^{(k)}\beta-u - \alpha\|_2^2 + \lambda/\rho
\|D^{(1)}\alpha\|_1,\\
\label{eq:u2}
u &\leftarrow u + \alpha - D^{(k)}\beta.
\end{align}
The $\beta$- and $u$-updates are analogous to those from the standard
ADMM, just of a smaller order $k$. But the $\alpha$-update above is
not; the $\alpha$-update itself requires solving a constant order
trend filtering problem, i.e., a 1d fused lasso problem.  Therefore,
the specialized approach would not be efficient if it were not for the
extremely fast, direct solvers that exist for the 1d fused lasso.  Two
exact, linear time 1d fused lasso solvers were given by
\citet{tautstring}, \citet{nickdp}.  The former is based on taut
strings, and the latter on dynamic programming.  Both algorithms are
very creative and are a marvel in their own right; we are more
familiar with the dynamic programming approach, and so in our
specialized ADMM algorithm, we utilize (a custom-made,
highly-optimized implementation of) this dynamic programming routine
for the $\alpha$-update, hence writing  
\begin{equation}
\label{eq:dp}
\alpha \leftarrow \DP_{\lambda/\rho}(D^{(k)}\beta - u).
\end{equation}
This uses $O(n-k)$ operations, and thus a full round of specialized
ADMM updates runs in linear time, the same as the standard
ADMM ones (the two approaches are also empirically very similar in
terms of computational time; see Figure \ref{fig:runtimes}).  As mentioned in the introduction, neither the taut string nor dynamic
programming approach can be directly extended beyond the $k=0$ case,
to the best of our knowledge, for solving higher order trend filtering 
problems; however, they can be wrapped up in the special ADMM 
algorithm described above, and in this manner, they lend their
efficiency to the computation of higher order estimates.

\subsection{Superiority of specialized over standard ADMM}  

We now provide further experimental evidence that our specialized
ADMM implementation significantly outperforms the naive standard ADMM.   
We simulated noisy data from three different underlying signals: 
constant, sinusoidal, and Doppler wave signals (representing three
broad classes of functions: trivial smoothness, homogeneous
smoothness, and inhomogeneous smoothness).
We examined 9 different
problem sizes, spaced roughly logarithmically from $n=500$ to 
$n=500,000$, and considered computation of the trend 
filtering solution in \eqref{eq:tf} for the orders $k=1,2,3$. 
We also considered 20 values of $\lambda$, spaced
logarithmically between $\lambda_{\max}$ and $10^{-5} \lambda_{\max}$, 
where 
\begin{equation*}
\lambda_{\max} = \big\|\big((D^{(k+1)} (D^{(k+1)})^T\big)^{-1
}(D^{(k+1)})^T y\big\|_\infty,
\end{equation*}
the smallest value of $\lambda$ at which the penalty term 
\smash{$\|D^{(k+1)}\hbeta\|_1$} is zero at the solution (and hence the
solution is exactly a $k$th order polynomial).  In each problem
instance---indexed by the choice of underlying function, problem size,
polynomial order $k$, and tuning parameter value $\lambda$---we
ran a large number of iterations of the ADMM algorithms, and recorded 
the achieved criterion values across iterations.

\begin{figure}[h!]
\centering
\includegraphics[width=0.32\textwidth]{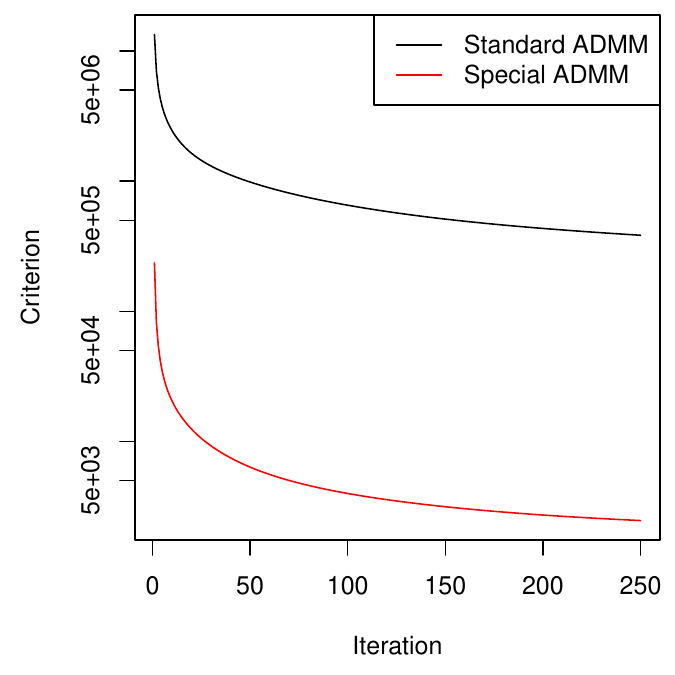} 
\includegraphics[width=0.32\textwidth]{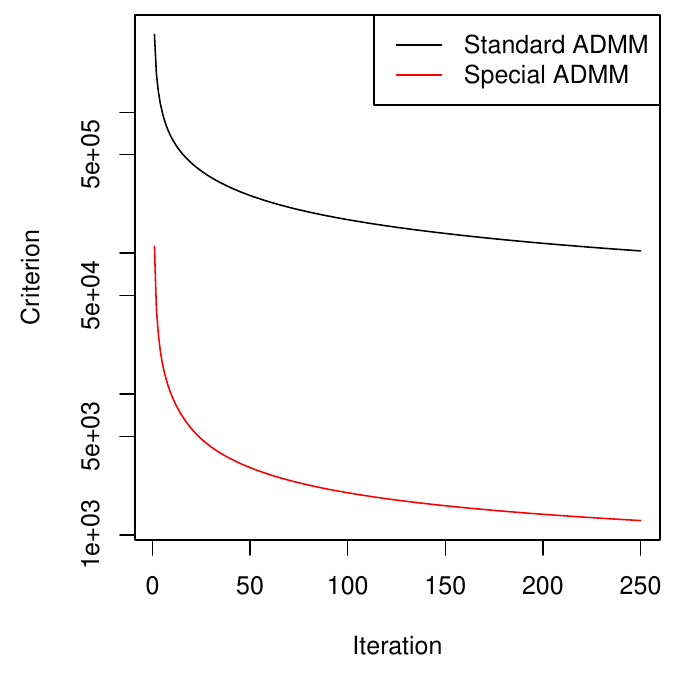}
\includegraphics[width=0.32\textwidth]{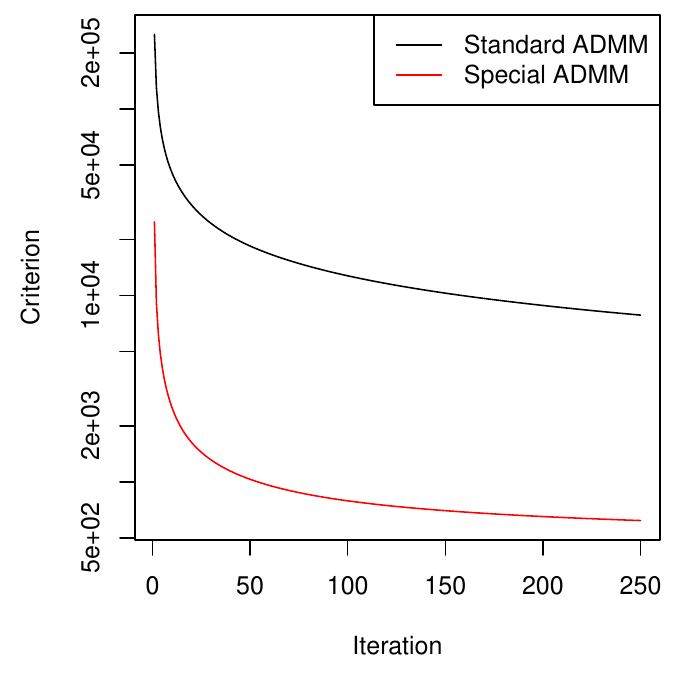}
\caption{\small\it All plots show values of the trend filtering 
 criterion versus iteration number in the two ADMM implementations.  
  The underlying signal here was the Doppler wave, with $n=10,000$,
  and $k=2$.  The left plot shows a large value of $\lambda$ (near
  $\lambda_{\max}$), the middle a medium value (halfway in between
  $\lambda_{\max}$ and $10^{-5}\lambda_{\max}$, on a log scale), and
  the right a small value (equal to $10^{-5}\lambda_{\max}$).  The
  specialized ADMM approach easily outperforms the standard one in all
  cases.} 
\label{fig:stdVSspl} 
\end{figure}

The results from one particular instance, 
in which the underlying signal was the Doppler wave, $n=10,000$, and
$k=2$, are shown in Figure \ref{fig:stdVSspl}; this instance was
chosen arbitrarily, and we have found the same qualitative behavior to  
persist throughout the entire simulation suite.  
We can see clearly that in each regime 
of regularization, the specialized routine dominates the standard one in terms of
convergence to optimum.  Again, we reiterate that qualitatively 
the same conclusion holds across all simulation parameters, and
the gap between the specialized and standard approaches generally
widens as the polynomial order $k$ increases.

\subsection{Some intuition for specialized versus standard ADMM} 
\label{sec:intuition1}

One may wonder why the two algorithms, standard and specialized ADMM,   
differ so significantly in terms of their performance.  Here we
provide some intuition with regard to this question.  A first, very
rough interpretation: the specialized algorithm utilizes a dynamic
programming subroutine \eqref{eq:dp} in place of soft-thresholding
\eqref{eq:soft}, therefore solving a more ``difficult'' subproblem in
the same amount of time (linear in the input size),
and likely making more progress towards minimizing the overall
criterion.  In other words, this reasoning follows the underlying 
intuitive principle that for a given optimization task, an ADMM
parametrization with ``harder'' subproblems will enjoy faster  
convergence.   

While the above explanation was fairly vague,
a second, more concrete explanation comes from viewing the 
two ADMM routines in ``basis'' form, i.e., from essentially inverting 
$D^{(k+1)}$ to yield an equivalent lasso form of trend filtering, as
explained in \eqref{eq:lasso} of Appendix \ref{app:motivate}, where
$H^{(k)}$ is a basis matrix.  From this equivalent perspective, the
standard ADMM algorithm reparametrizes \eqref{eq:lasso} as in
\begin{equation}
\label{eq:lasso1}
\min_{\theta \in \R^n, \, w \in \R^n} \, \half 
\|y - H^{(k)} w\|_2^2 + \lambda \cdot k!
\sum_{j=k+2}^n |\theta_j| \;\;\st\;\; w = \theta,
\end{equation}
and the specialized ADMM algorithm reparametrizes \eqref{eq:lasso} as
in 
\begin{equation}
\label{eq:lasso2}
\min_{\theta \in \R^n, \, w \in \R^n} \, \half 
\|y - H^{(k-1)} w\|_2^2 + \lambda \cdot k!
\sum_{j=k+2}^n |\theta_j| \;\;\st\;\; w = L \theta,
\end{equation}
where we have used the recursion $H^{(k)}=H^{(k-1)} L$
\citep{fallfact}, analogous (equivalent) to $D^{(k+1)} = D^{(1)}
D^{(k)}$.
The matrix $L \in \R^{n\times n}$ is block diagonal with the
first $k \times k$ block being the identity, and the last
$(n-k)\times (n-k)$ block being the lower triangular matrix of 1s.
What is so different between applying ADMM to \eqref{eq:lasso2}
instead of \eqref{eq:lasso1}?  Loosely speaking, if we ignore the role
of the dual variable, the ADMM steps can be thought of as performing
alternating minimization over $\theta$ and $w$. The joint criterion
being minimized, i.e., the augmented Lagrangian (again hiding the
dual variable) is of the form
\begin{equation}
\label{eq:lag1}
\half \left\| z - 
\left[\begin{array}{cc}
H^{(k)} & 0 \\
\sqrt{\rho}I & -\sqrt{\rho}I
\end{array}\right] 
\left[\begin{array}{cc}
\theta \\ w
\end{array}\right]
\right\|_2^2 + \lambda \cdot k!
\sum_{j=k+2}^n |\theta_j|
\end{equation}
for the standard parametrization \eqref{eq:lasso1}, and 
\begin{equation}
\label{eq:lag2}
\half \left\| z - 
\left[\begin{array}{cc}
H^{(k-1)} & 0 \\
\sqrt{\rho}I & -\sqrt{\rho}L
\end{array}\right] 
\left[\begin{array}{cc}
\theta \\ w
\end{array}\right]
\right\|_2^2 + \lambda \cdot k!
\sum_{j=k+2}^n |\theta_j|
\end{equation}
for the special parametrization \eqref{eq:lasso2}.  The key difference
between \eqref{eq:lag1} and \eqref{eq:lag2} is that the left and right
blocks of the regression matrix in \eqref{eq:lag1} are highly
(negatively) correlated (the bottom left and right blocks are each
scalar multiples of the identity), but the blocks of the regression
matrix in \eqref{eq:lag2} are not (the bottom blocks are the identity
and the lower triangular matrix of 1s).  Hence, in the 
context of an alternating minimization scheme, an update
step in \eqref{eq:lag2} should make more progress than an update step
in \eqref{eq:lag1}, because the descent directions for $\theta$ and
$w$ are not as adversely aligned (think of coordinatewise
minimization over a function whose contours are tilted ellipses,
and over one whose contours are spherical). Using the equivalence
between the basis form and the original (difference-penalized) form of
trend filtering, therefore, we may view the special ADMM updates
\eqref{eq:beta2}--\eqref{eq:u2} as {\it decorrelated} versions of the
original ADMM updates \eqref{eq:beta1}--\eqref{eq:u1}.  This allows 
each update step to make greater progress in descending on the overall 
criterion.

\subsection{Superiority of warm over cold starts}

\begin{figure}[h]
\centering
\includegraphics[width=0.48\textwidth]{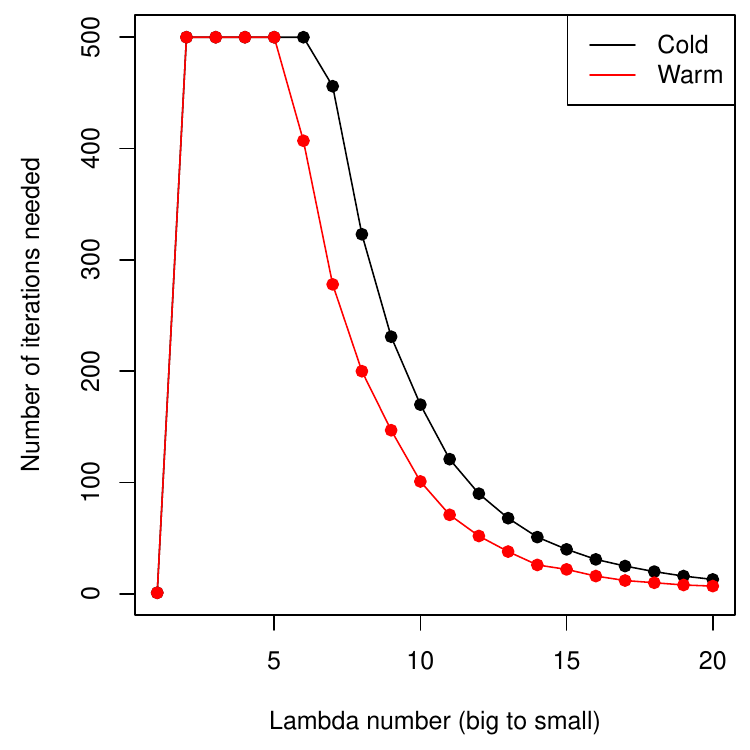}
\includegraphics[width=0.48\textwidth]{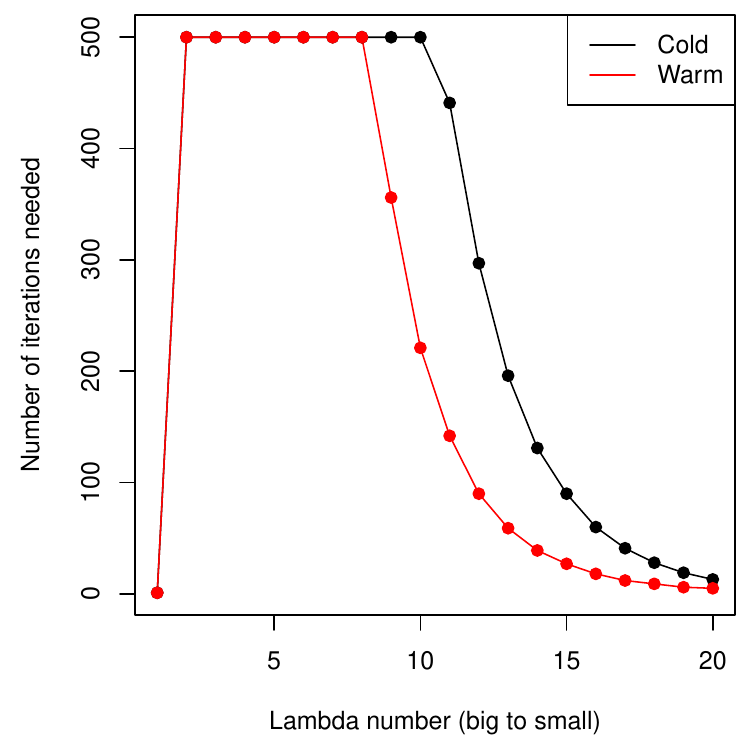}
\caption{\small\it The x-axis in both panels represents 20 values of 
  $\lambda$, log-spaced between $\lambda_{\max}$ and
  $10^{-5}\lambda_{\max}$, and the y-axis the number of iterations
  needed by specialized ADMM  to reach a prespecified
  level of accuracy, for $n=10,000$ noisy points from the Doppler
  curve for $k=2$ (left) and $k=3$ (right). Warm starts (red) have an
  advantage over cold starts (black), especially in the statistically
  reasonable (middle) range for $\lambda$.} 
\label{fig:warmVScold}
\end{figure}

In the above numerical comparison between special and standard ADMM,
we ran both methods with cold starts, meaning that the problems over
the sequence of $\lambda$ values were solved independently, without
sharing information.  Warm starting refers to a strategy in which we
solve the problem for the largest value of $\lambda$ first, use this
solution to initialize the algorithm at the second largest value of
$\lambda$, etc. With warm starts, the relative performance of the two
ADMM approaches does not change. However, the performance of both
algorithms does improve in an absolute sense,  illustrated for the
specialized ADMM algorithm in Figure \ref{fig:warmVScold}. \\

This example is again representative of the experiments
across the full simulation suite.  Therefore, from this point forward,
we use warm starts for all experiments. 


\subsection{Choice of the augmented Lagrangian parameter $\rho$} 

A point worth discussing is the choice of augmented Lagrangian
parameter $\rho$ used in the above experiments. Recall that $\rho$ is
not a statistical parameter associated with the trend filtering
problem \eqref{eq:tf}; it is rather an optimization parameter
introduced during the formation of the agumented Lagrangian in ADMM.
It is known that under very general conditions, the ADMM algorithm
converges to optimum for any fixed value of $\rho$ \citep{admm};
however, in practice, the rate of convergence of the algorithm, as
well as its numerical stability, can both depend strongly on the
choice of $\rho$. 

We found the choice of setting $\rho=\lambda$ to be numerically stable across all
setups. Note that in the ADMM updates \eqref{eq:beta1}--\eqref{eq:u1}
or  \eqref{eq:beta2}--\eqref{eq:u2}, the only appearance of $\lambda$
is in the $\alpha$-update, where we apply $S_{\lambda/\rho}$ or  
$\DP_{\lambda/\rho}$, soft-thresholding or dynamic programming (to
solve the 1d fused lasso problem) at the level $\lambda/\rho$.
Choosing $\rho$ to be proportional to $\lambda$ controls the amount of
change enacted by these subroutines (intuitively making it neither too large nor too small at each step). 
We also tried adaptively varying $\rho$, a heuristic suggested by  
\citet{admm}, but found this strategy to be less stable overall; it
did not yield consistent benefits for either algorithm.

Recall that this paper is not concerned with the model selection problem of how to choose $\lambda$, but just with the optimization problem of how to solve \eqref{eq:tf} when given $\lambda$. All results in the rest of this paper reflect the default choice
$\rho=\lambda$, unless stated otherwise.

\section{Comparison of specialized ADMM and PDIP} 
\label{sec:compare}

Here we compare our specialized ADMM algorithm and the PDIP algorithm 
of \citep{l1tf}. We used the C++/LAPACK
implementation of the PDIP method (written for the case $k=1$) that is 
provided freely by these authors, and generalized it to work for an
arbitrary order $k \geq 1$.  To put the methods on equal 
footing, we also wrote our own efficient C implementation of
the specialized ADMM algorithm.
This code has been interfaced to R via the {\tt trendfilter} function in
the R package {\tt glmgen}, available at 
\url{https://github.com/statsmaths/glmgen}.

A note on the PDIP implementation: 
this algorithm is actually applied to the dual of
\eqref{eq:tf}, as given in \eqref{eq:dual} in Appendix
\ref{app:motivate}, and its iterations solve linear systems
  in the banded matrix $D$ in $O(n)$ time. 
The number of constraints, and hence the number of log barrier terms,
is $2(n-k-1)$.  We used 10 for the log barrier update factor
(i.e., at each iteration, the weight of log barrier term is scaled by
$1/10$).  We used backtracking line search to choose the step size in 
each iteration, with parameters 0.01 and 0.5 (the former being the
fraction of improvement over the gradient required to exit, and the
latter the step size shrinkage factor).  These specific parameter
values are the defaults suggested by \citet{convex} for interior
point methods, and are very close to the defaults in the original PDIP
linear trend filtering code from \citet{l1tf}. In the settings in
which PDIP struggled (to be seen in what follows), we tried varying
these parameter values, but no single choice led to consistently
improved performance.   

\subsection{Comparison of cost per iteration}

\begin{figure}[h]
\centering
\includegraphics[width=0.55\textwidth]{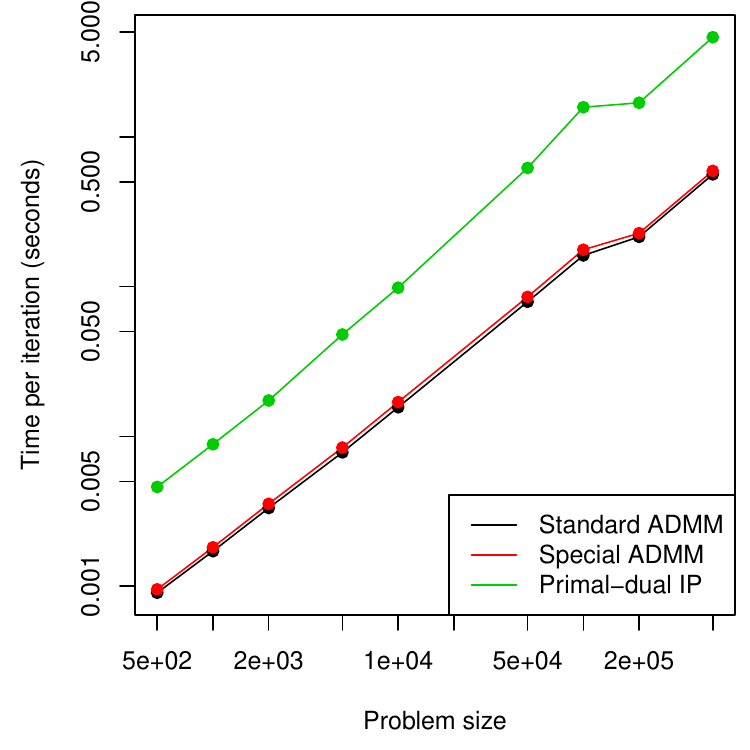}
\caption{\small\it A log-log plot of time per iteration of ADMM
  and PDIP routines against  problem size $n$ (20 values from 500
  up to 500,000). The times per iteration of the algorithms were
  averaged over 3 choices of underlying function (constant,
  sinusoidal, and Doppler), 3 orders of trends ($k=1,2,3$), 20 values 
  of $\lambda$ (log-spaced between $\lambda_{\max}$ and
  $10^{-5}\lambda_{\max}$), and 10 repetitions for each combination
  (except the two largest problem sizes, for which we performed 3
  repetitions).  This validates the theoretical $O(n)$
  iteration complexities of the algorithms, and the larger offset (on
  the log-log scale) for PDIP versus ADMM implies a larger
  constant in the linear scaling: an ADMM iteration is about 
  10 times faster than a PDIP iteration.}  
\label{fig:runtimes}
\end{figure}

Per iteration, both ADMM and PDIP take
$O(n)$ time, as explained earlier.  Figure
\ref{fig:runtimes} reveals that the constant
hidden in the $O(\cdot)$ notation is about 10 times larger for PDIP
than ADMM. Though the comparisons that follow are based on achieved
criterion value versus iteration, it may be kept in mind that
convergence plots for the criterion values versus time would be
stretched by a factor of 10 for PDIP.     \\

\subsection{Comparison for $k=1$ (piecewise linear fitting)}

\begin{figure}[p]
\centering

\begin{subfigure}{\textwidth}
\centering
\includegraphics[width=0.32\textwidth]{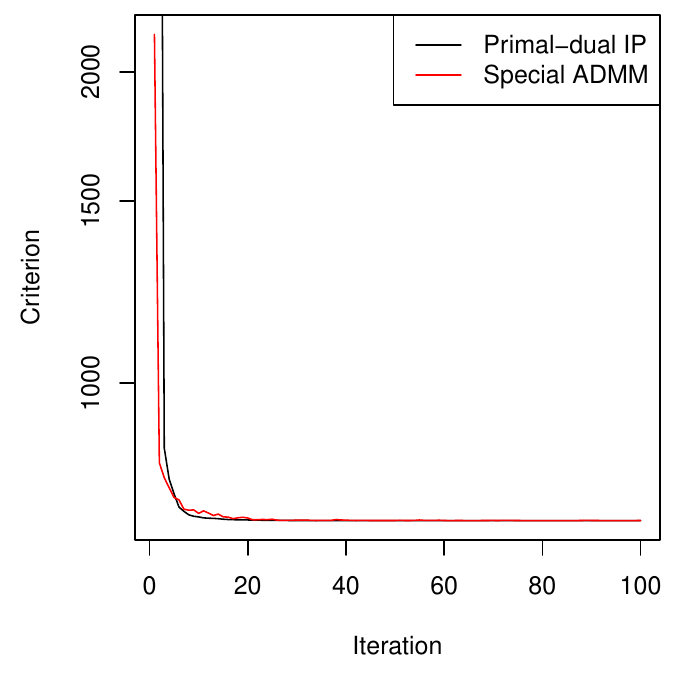} 
\includegraphics[width=0.32\textwidth]{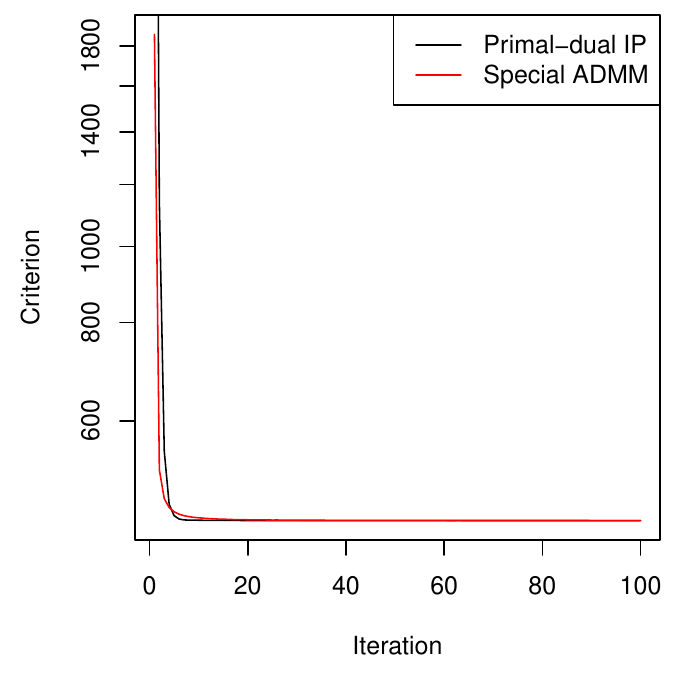}
\includegraphics[width=0.32\textwidth]{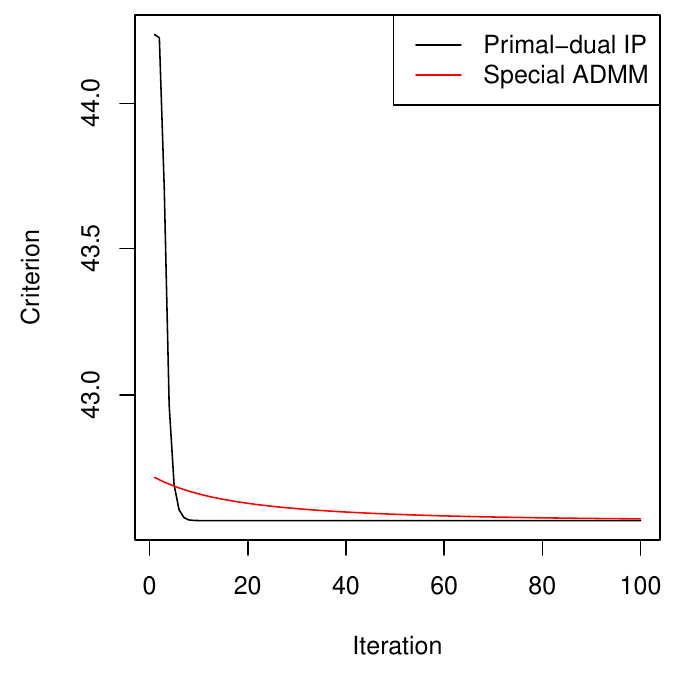} \\
\includegraphics[width=0.32\textwidth]{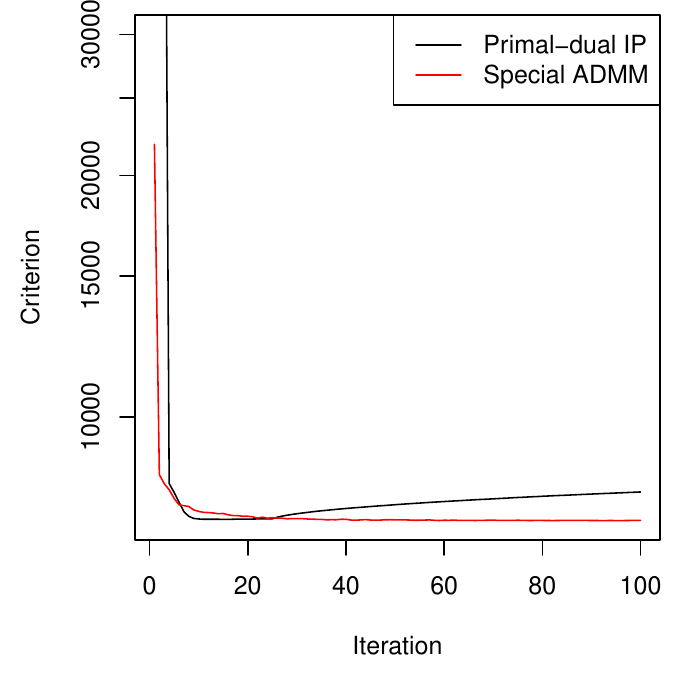} 
\includegraphics[width=0.32\textwidth]{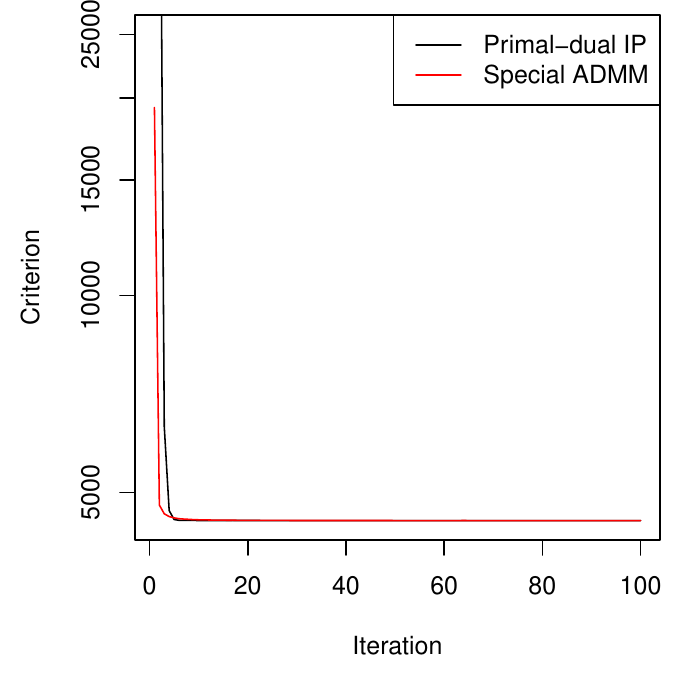}
\includegraphics[width=0.32\textwidth]{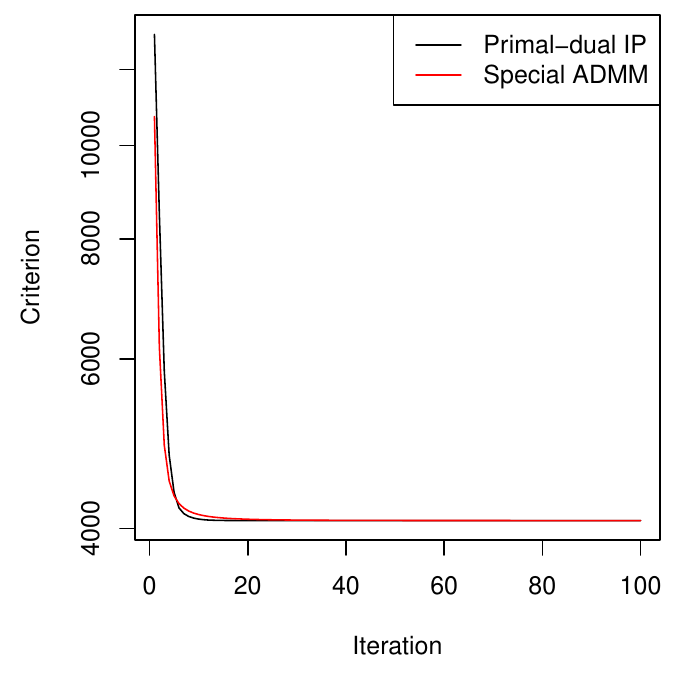}
\caption{\small\it Convergence plots for $k=1$: achieved criterion
  values across iterations of ADMM and PDIP.  The first row
  concerns $n=10,000$ points, and the second row $n=100,000$ points,
  both generated around a sinusoidal 
  curve. The columns (from left to right) display high to low values
  of $\lambda$: near $\lambda_{\max}$, halfway in between (on a log 
  scale) $\lambda_{\max}$ and $10^{-5}\lambda_{\max}$, and equal to 
  $10^{-5}\lambda_{\max}$, respectively. Both algorithms exhibit good 
  convergence.}   
\label{fig:crit-pd-admm-k1-conv}
\end{subfigure}

\bigskip
\begin{subfigure}{\textwidth}
\centering
\includegraphics[width=0.24\textwidth]{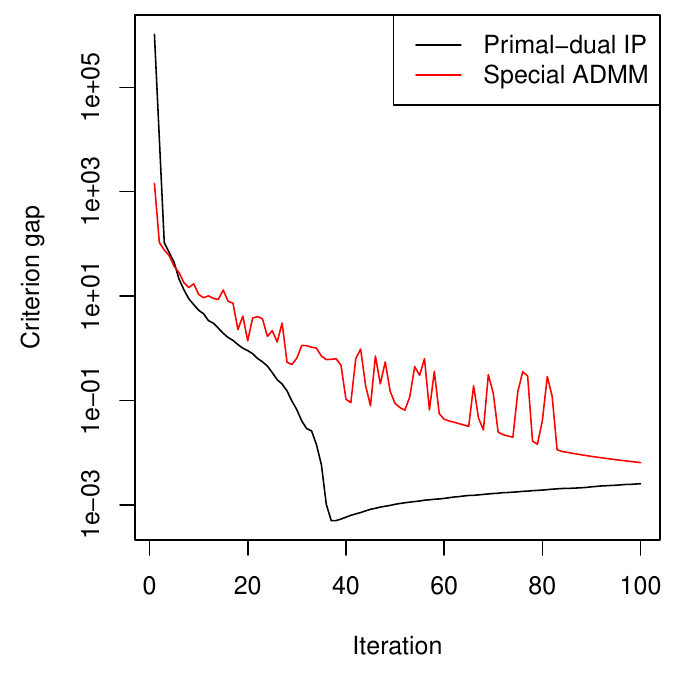} 
\includegraphics[width=0.24\textwidth]{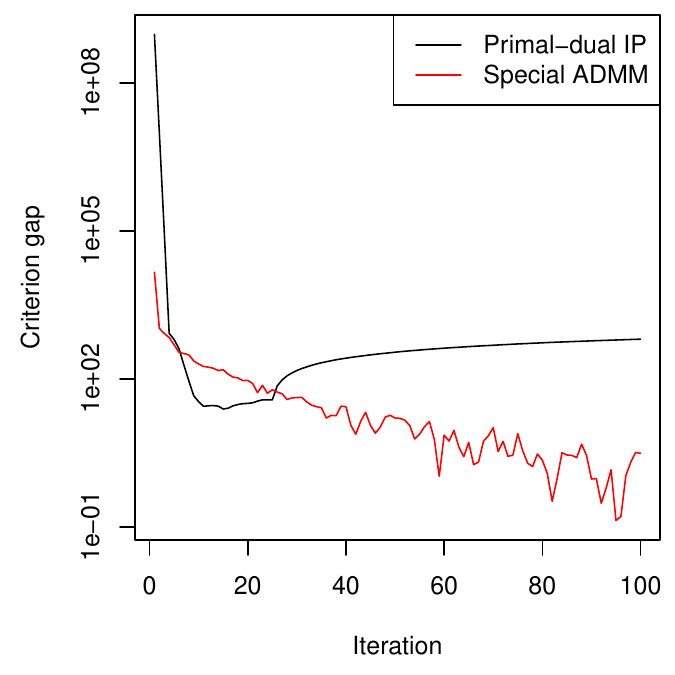} 
\includegraphics[width=0.24\textwidth]{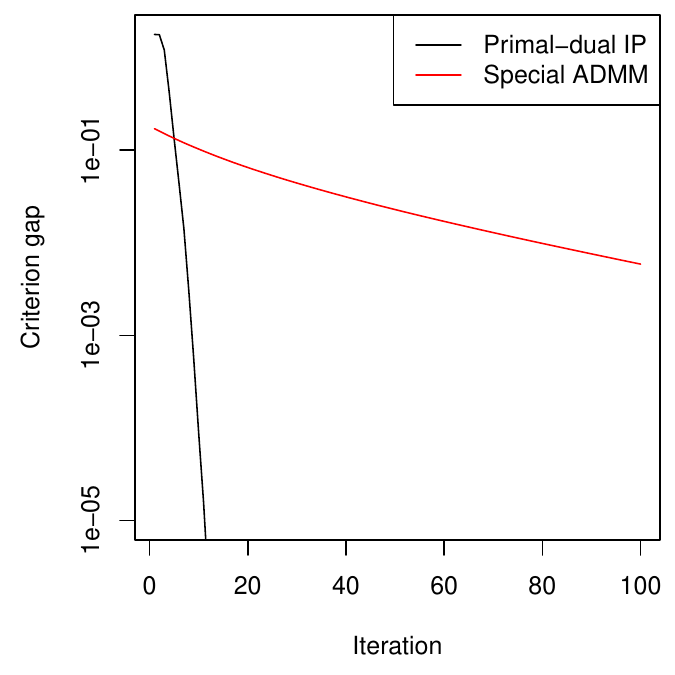}
\includegraphics[width=0.24\textwidth]{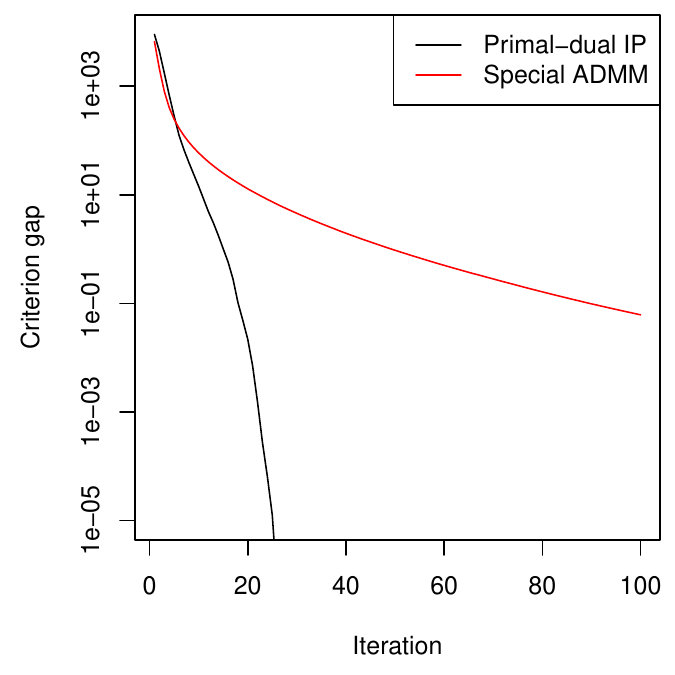}
\caption{\small\it Convergence gaps for $k=1$: achieved criterion
  value minus the optimum value across iterations of ADMM 
  and PDIP.  Here the optimum value was defined as smallest achieved
  criterion value over 5000 iterations of either algorithm.  The first
  two plots are for $\lambda$ near $\lambda_{\max}$, with $n=10,000$ 
  and $n=100,000$ points, respectively.  In this high regularization
  regime, ADMM fares better for large $n$. The last two plots are for 
  $\lambda=10^{-5}\lambda_{\max}$, with $n=10,000$ and $n=100,000$, 
  respectively.  Now in this low regularization regime, PDIP converges
  at what appears to be a second-order rate, and ADMM does
  not. However, these small values of $\lambda$ are not statistically
  interesting in the context of the example, as they yield grossly
  overfit trend estimates of the underlying sinusoidal curve.}
\label{fig:crit-pd-admm-k1-gap}
\end{subfigure}

\caption{\small\it Convergence plots and gaps ($k=1$), 
  for specialized ADMM and PDIP.}
\label{fig:crit-pd-admm-k1}
\end{figure}

In general, for $k=1$ (piecewise linear fitting), both the specialized
ADMM and PDIP algorithms perform similarly, as displayed in Figure
\ref{fig:crit-pd-admm-k1}. The PDIP algorithm displays a relative
advantage as $\lambda$ becomes small, but the convergence of ADMM is
still strong in absolute terms. Also, it is important to note that
these small values of $\lambda$ correspond to solutions that overfit
the underlying trend in the problem context, and hence PDIP
outperforms ADMM in a statistically uninteresting regime of
regularization.  

\subsection{Comparison for $k=2$ (piecewise quadratic fitting)} 

\begin{figure}[p]
\centering

\begin{subfigure}{\textwidth}
\centering
\includegraphics[width=0.32\textwidth]{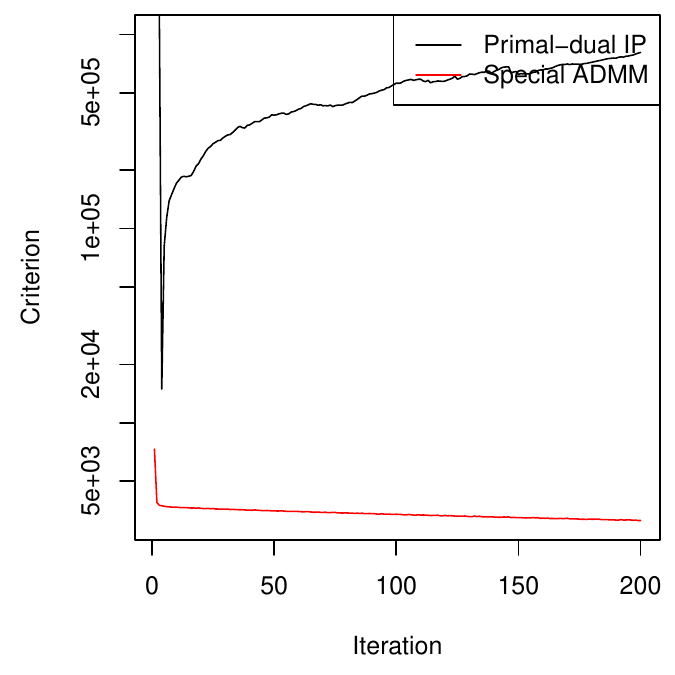} 
\includegraphics[width=0.32\textwidth]{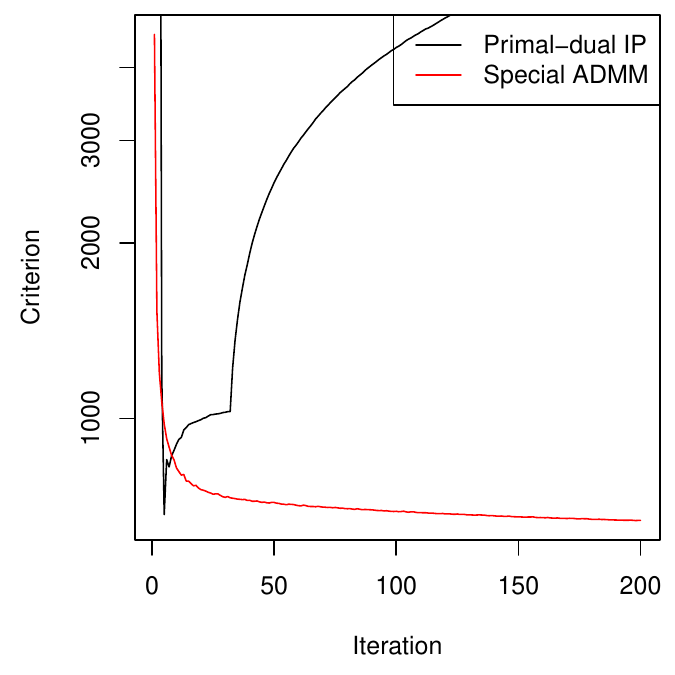}
\includegraphics[width=0.32\textwidth]{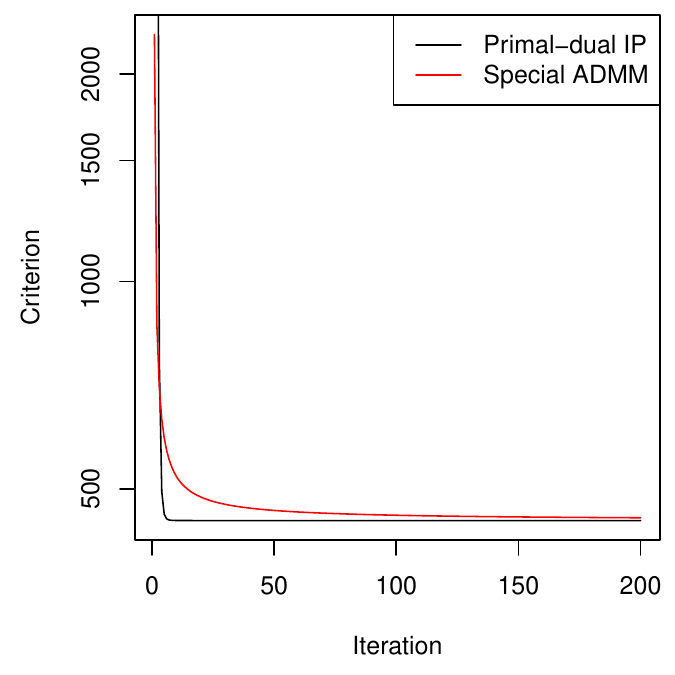} \\
\includegraphics[width=0.32\textwidth]{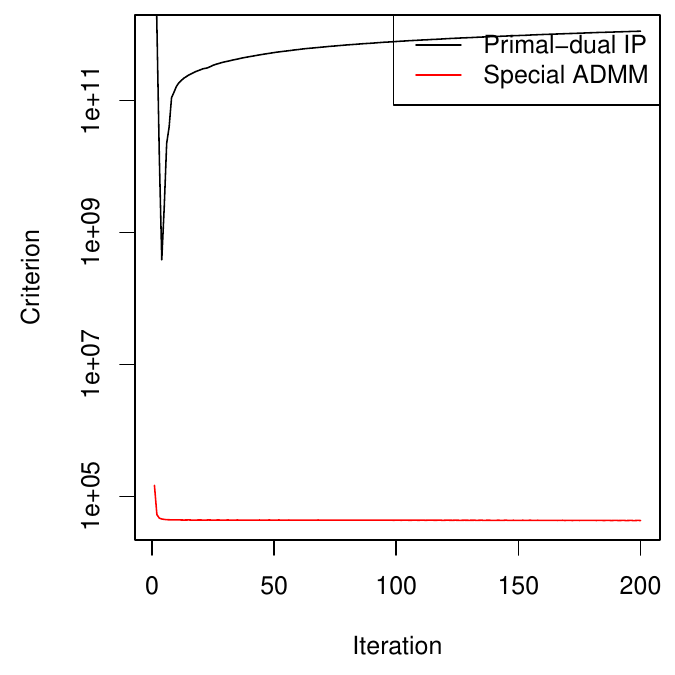} 
\includegraphics[width=0.32\textwidth]{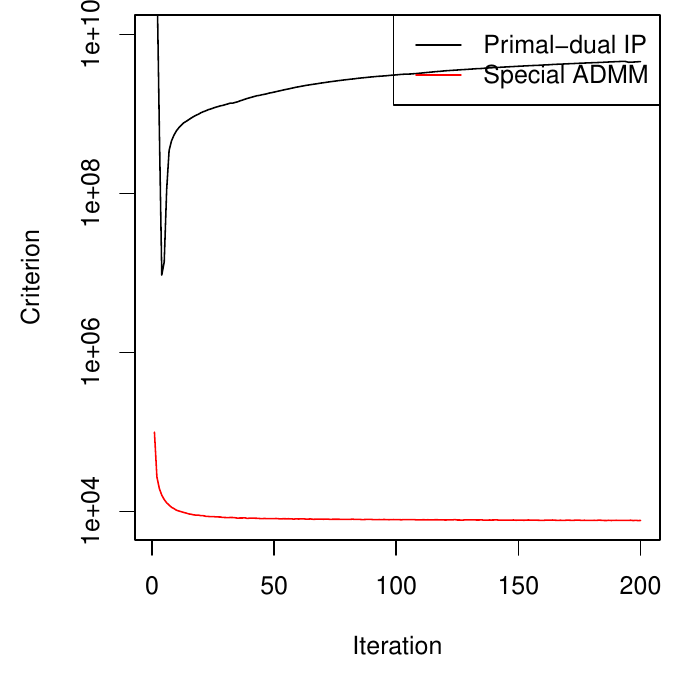}
\includegraphics[width=0.32\textwidth]{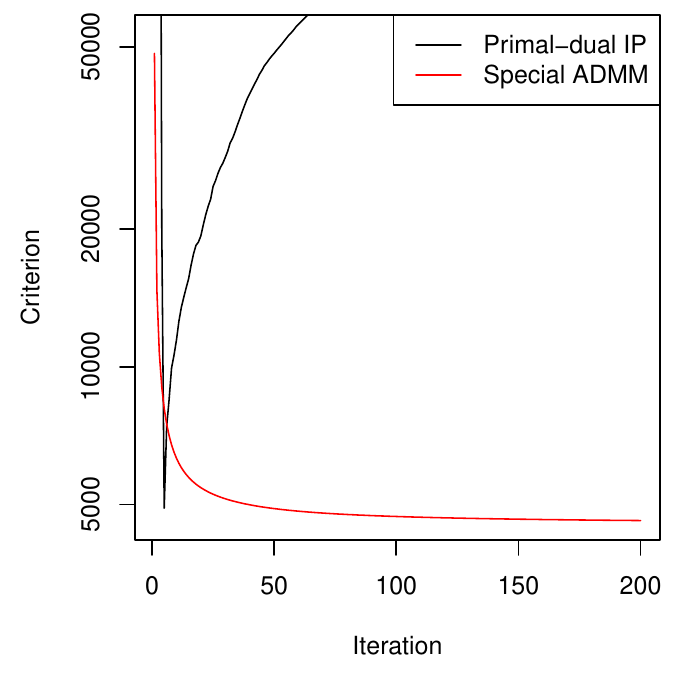}
\caption{\small\it Convergence plots for $k=2$: achieved criterion 
  values across iterations of ADMM and PDIP, 
  with the same layout as in Figure \ref{fig:crit-pd-admm-k1-conv}.
  The specialized ADMM routine has fast convergence in all cases.
  For all but the smallest $\lambda$ values, PDIP does
  not come close to convergence.  These values of $\lambda$ are so
  small that the corresponding trend filtering solutions are not
  statistically desirable in the first place; see below.}
\label{fig:crit-pd-admm-k2-conv}
\end{subfigure}

\bigskip
\begin{subfigure}{\textwidth}
\centering
\includegraphics[width=0.24\textwidth]{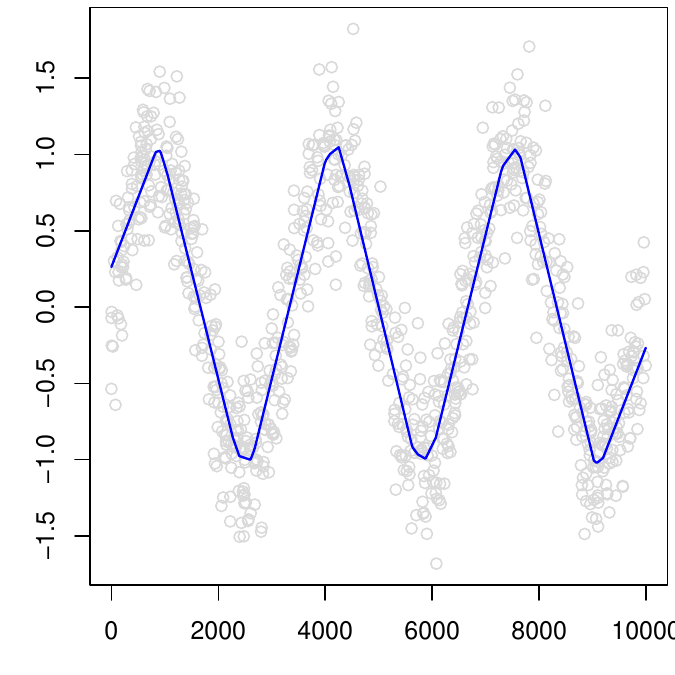} 
\includegraphics[width=0.24\textwidth]{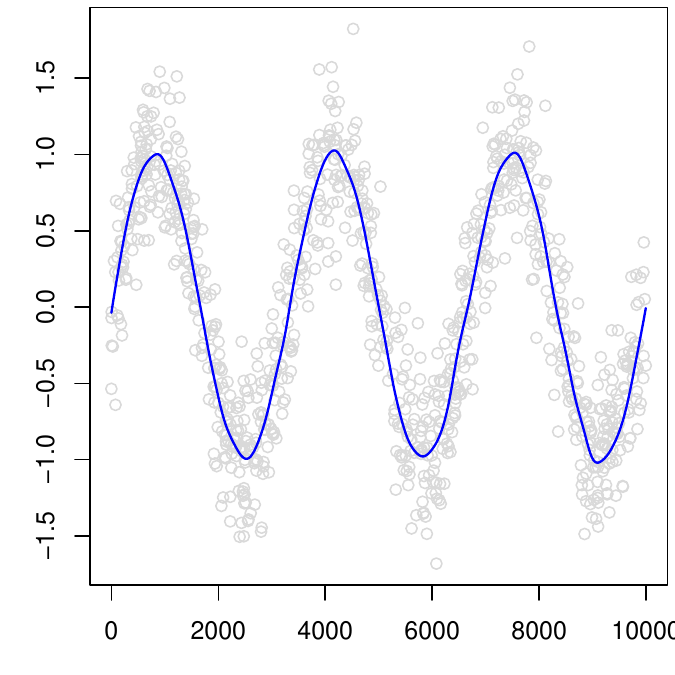}
\includegraphics[width=0.24\textwidth]{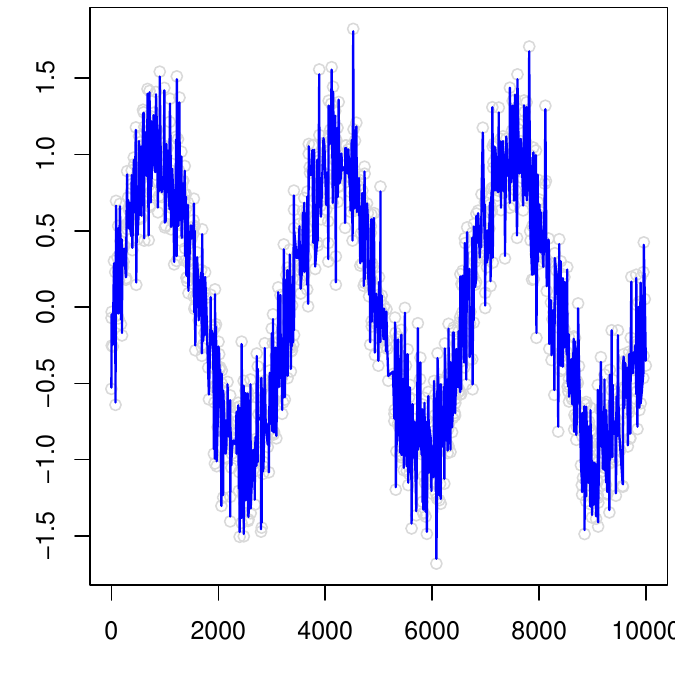} 
\includegraphics[width=0.24\textwidth]{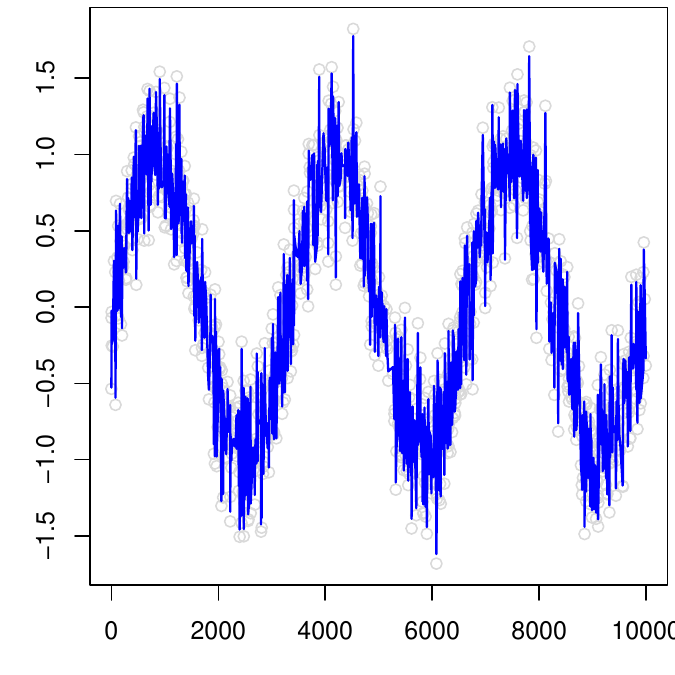}
\caption{\small\it 
  Visualization of trend filtering estimates for the 
  experiments in Figures \ref{fig:crit-pd-admm-k1}, 
  \ref{fig:crit-pd-admm-k2}.  The estimates were trained on
  $n=10,000$ points from an underlying sinusoidal curve
  (but the above plots have been downsampled to 1000 points for 
  visibility).  The two left panels show the fits
  for $k=1,2$, respectively, in the high regularization regime,
  where $\lambda$ is near $\lambda_{\max}$. The
  specialized ADMM approach outperforms PDIP (and shown are the ADMM
  fits).  The two right panels show the fits for $k=1,2$,
  respectively, in the low regularization regime, with 
  $\lambda=10^{-5}\lambda_{\max}$. 
  PDIP converges faster than ADMM (and shown are the PDIP fits),
  but this is not a statistically reasonable regime for trend 
  estimation.}  
\label{fig:trend-pd-admm-k2}
\end{subfigure}

\caption{\small\it Convergence plots and estimated fits ($k=2$) for
  special ADMM and PDIP.}  
\label{fig:crit-pd-admm-k2}
\end{figure}

For $k=2$ (piecewise quadratic fitting), the PDIP routine struggles
for moderate to large values of $\lambda$, increasingly so as the
problem size grows, as shown in Figure \ref{fig:crit-pd-admm-k2}.
These convergence issues remain as we vary its internal optimization
parameters (i.e., its log barrier update parameter, and backtracking
parameters).  Meanwhile, our specialized ADMM approach is much more
stable, exhibiting strong convergence behavior across all $\lambda$
values, even for large problem sizes in the hundreds of thousands. 

The convergence issues encountered by PDIP here, when $k=2$, are 
only amplified when $k=3$, as the issues begin to show at much smaller
problem sizes; still, the specialized ADMM steadily converges, and is
a clear winner in terms of robustness. Analysis of this case is
deferred until Appendix \ref{app:k3} for brevity.

\subsection{Some intuition on specialized ADMM versus PDIP}
\label{sec:intuition2}

We now discuss some intuition for the observed
differences between the specialized ADMM and PDIP. 
This experiments in this section showed that PDIP will often diverge
for large problem sizes and moderate values of the trend order 
($k=2,3$), regardless of the choices of the log barrier and
backtracking line search parameters.  That such behavior presents
itself for large $n$ and $k$ suggests that PDIP is affected by poor
conditioning of the difference operator $D^{(k+1)}$ in these cases.  
Since PDIP is affine invariant, in theory it should not be affected by
issues of conditioning at all.  But when $D^{(k+1)}$ is poorly
conditioned, it is difficult to solve the linear systems in
$D^{(k+1)}$ that lie at the core of a PDIP iteration, and this leads
PDIP to take a noisy update step (like taking a majorization 
step using a perturbed version of the Hessian).  If
the computed update directions are noisy enough, then PDIP can
surely diverge.   

Why does specialized ADMM  not suffer the same fate,
since it too solves linear systems in each iteration (albeit in
$D^{(k)}$ instead of $D^{(k+1)}$)?  There is an important difference
in the form of these linear systems.  Disregarding the order of
the difference operator and denoting it simply by $D$, a PDIP
iteration solves linear systems (in $x$) of the form 
\begin{equation}
\label{eq:lin1}
(DD^T + J)x = b
\end{equation}
where $J$ is a diagonal matrix, and an ADMM
iteration solves systems of the form
\begin{equation}
\label{eq:lin2}
(\rho D^T D + I) x = b
\end{equation}
The identity matrix $I$ provides an important eigenvalue
``buffer'' for the linear system in \eqref{eq:lin2}: the 
eigenvalues of $\rho D^T D + I$ are all bounded away from zero (by 1),
which helps make up for the poor conditioning inherent to
$D$. Meanwhile, the diagonal elements of $J$ in \eqref{eq:lin1} can be
driven to zero across iterations of the PDIP method; in fact, at
optimality, complementary slackness implies that $J_{ii}$ is zero
whenever the $i$th dual variable lies strictly inside the interval 
$[-\lambda,\lambda]$. Thus, the matrix $J$ does not always provide the 
needed buffer for the linear system in \eqref{eq:lin1}, so $DD^T + J$
can remain poorly conditioned, causing numerical instability issues
when solving \eqref{eq:lin1} in PDIP iterations.  In particular, when
$\lambda$ is large, many dual coordinates will lie strictly inside
$[-\lambda,\lambda]$ at optimality, which means that many diagonal
elements of $J$ will be pushed towards zero over PDIP iterations. 
This explains why PDIP experiences particular difficulty in the large
$\lambda$ regime, as  seen in our experiments.

\section{Arbitrary input points}
\label{sec:uneven}

Up until now, we have assumed that the input locations are implicitly 
$x_1=1, \ldots x_n=n$;  
in this section, we discuss the algorithmic extension of our
specialized ADMM algorithm to the case of arbitrary input points
$x_1,\ldots x_n$.  Such an extension is highly important, because, as a 
nonparametric regression tool, trend filtering is much more likely to
be used in a setting with generic inputs than one in
which these are evenly spaced.  Fortuitously, there is little that
needs to be changed with the trend filtering problem \eqref{eq:tf}
when we move from unit spaced inputs $1,\ldots n$ to arbitrary
ones $x_1,\ldots x_n$; the only difference is that the
operator $D^{(k+1)}$ is replaced by $D^{(x,k+1)}$, which is
adjusted for the uneven spacings present in $x_1,\ldots x_n$.  These 
adjusted difference operators are still banded with the same
structure, and are still defined recursively.  We begin
with $D^{(x,1)}=D^{(1)}$, the usual first difference operator in 
\eqref{eq:d1}, and then for $k\geq 1$, we define,
assuming unique sorted points $x_1 < \ldots < x_n$,
\begin{equation*}
D^{(x,k+1)} = D^{(1)} \cdot 
\diag\left( \frac{k}{x_{k+1}-x_1}, \ldots \frac{k}{x_n-x_{n-k}} \right)
\cdot D^{(x,k)},
\end{equation*}
where $\diag(a_1,\ldots a_m)$ denotes a diagonal matrix with 
 elements $a_1,\ldots a_m$; see \citet{trendfilter,fallfact}. 
Abbreviating this  
as \smash{$D^{(x,k+1)} = D^{(1)}\widetilde{D}^{(x,k)}$}, we see that
we only need to replace $D^{(k)}$ by \smash{$\widetilde{D}^{(x,k)}$} in
our special ADMM updates,
replacing one $(k+1)$-banded matrix with another. 


\begin{figure}[htbp]
\centering
\includegraphics[width=0.775\textwidth]{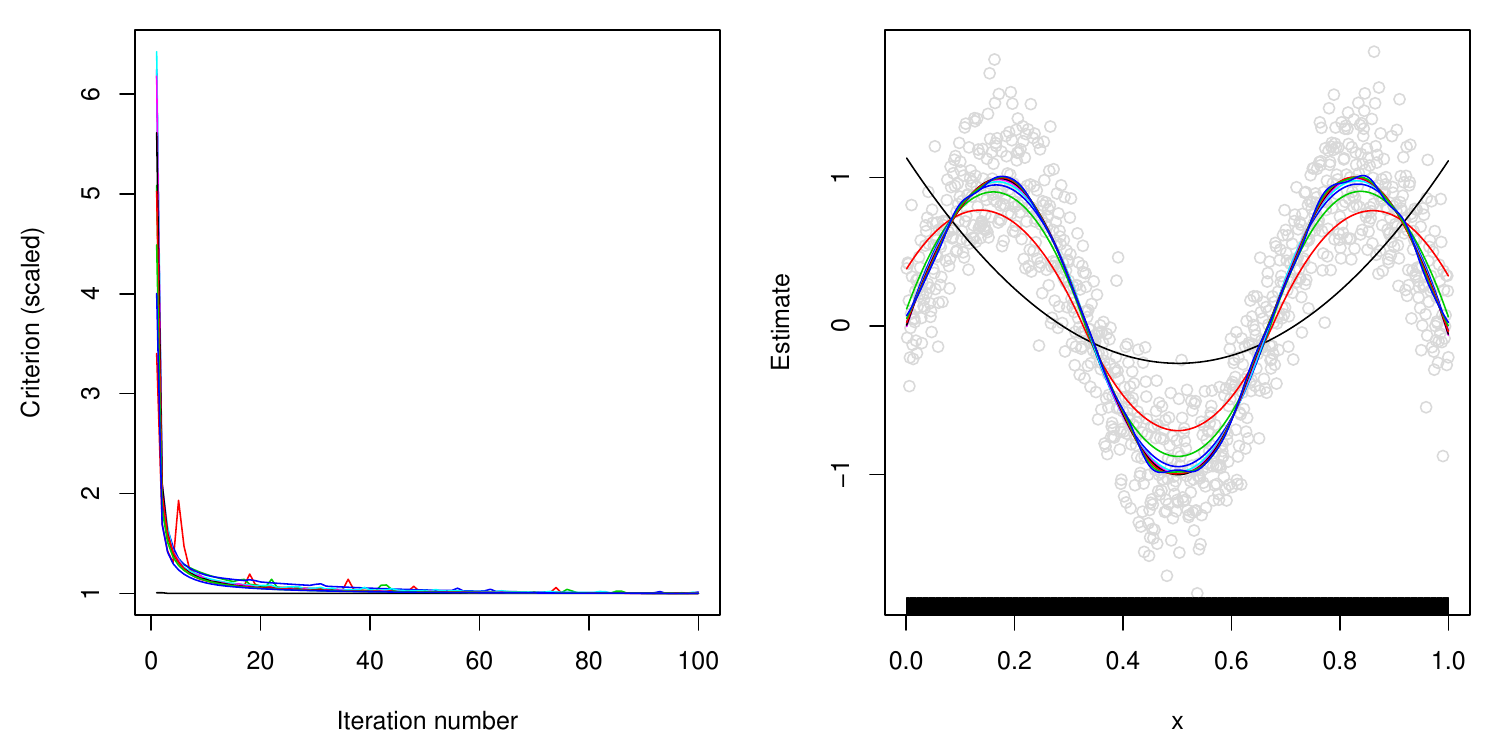} 
\includegraphics[width=0.775\textwidth]{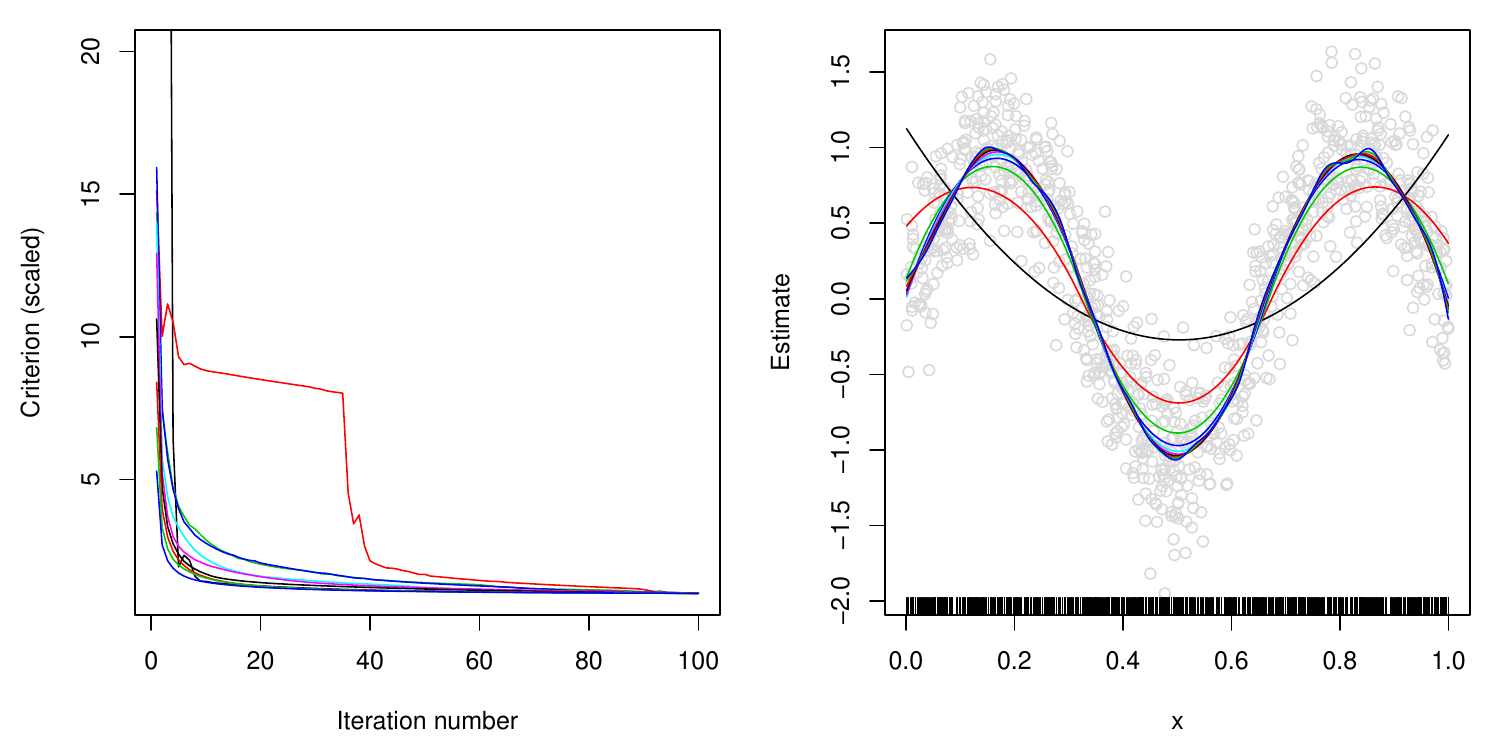}
\includegraphics[width=0.775\textwidth]{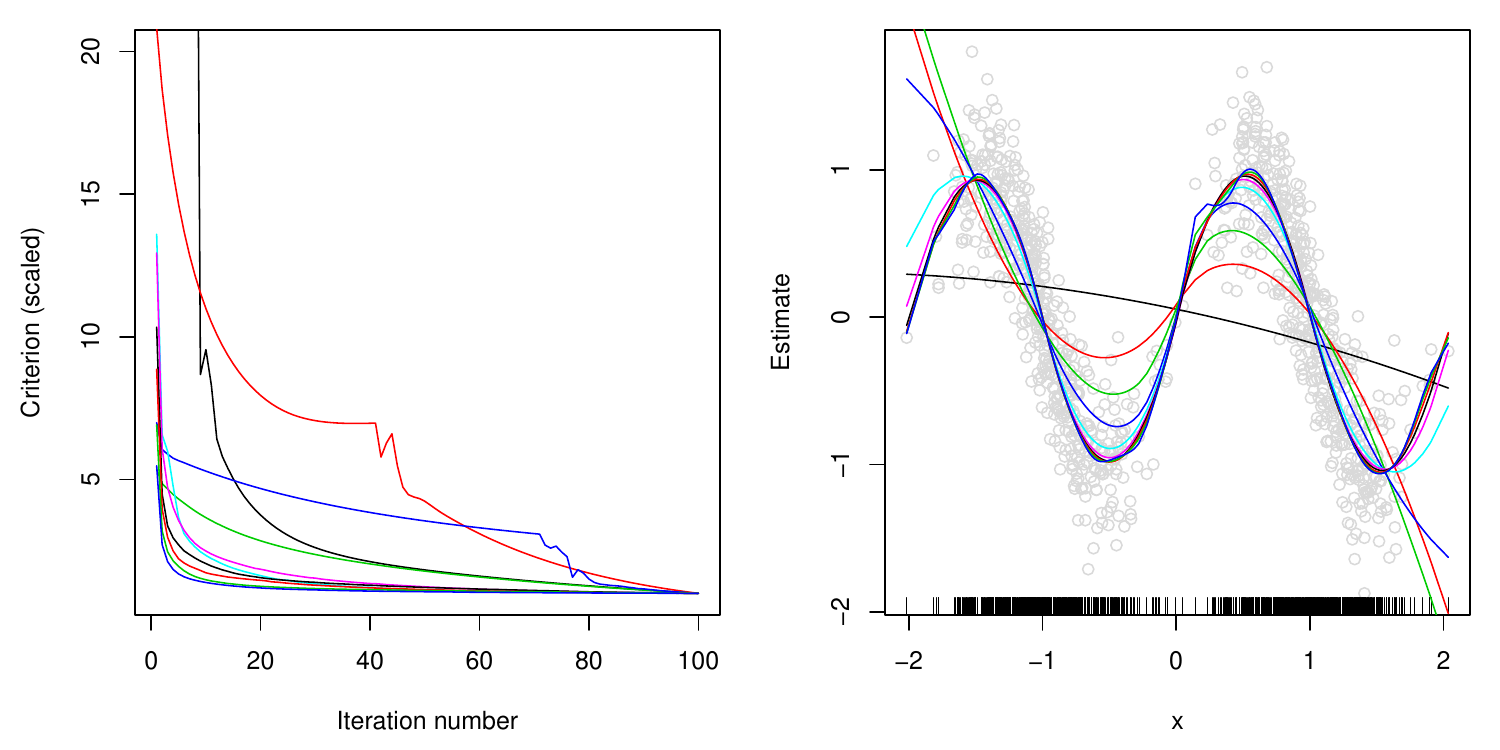} 
\caption{\it\small Each row considers a different design for the 
  inputs.  Top row: evenly spaced over $[0,1]$; middle row:
  uniformly at random over $[0,1]$; bottom row: mixture of Gaussians.
  In each case, we drew $n=1000$ points from a noisy sinusoidal curve  
  at the prescribed inputs.  The left panels show the
  achieved criterion values versus iterations of the specialized ADMM 
  implementation, with $k=2$, the different colored lines show
  convergence plots at different $\lambda$ values (we used 20 values  
  log-spaced between $\lambda_{\max}$ and $10^{-5}\lambda_{\max}$).
  The curves are all scaled to end at the same point for visibility.
  The ADMM algorithm experiences more difficulty as the input spacings
  become more irregular, due to poorer conditioning of the difference
  operator. The right panels plot the fitted estimates, with the ticks
  on the x-axis marking the input locations.}  
\label{fig:uneven}
\end{figure}

The more uneven the spacings among $x_1,\ldots x_n$, the worse the  
conditioning of \smash{$\tilde{D}^{(x,k)}$}, and hence the slower
to converge our specialized ADMM algorithm (indeed, the slower to
converge any of the alternative algorithms suggested in Section
\ref{sec:motivate}.)  As shown in
Figure \ref{fig:uneven}, however, our special ADMM approach is still
fairly robust even with considerably irregular design points
$x_1,\ldots x_n$.

\subsection{Choice of the augmented Lagrangian parameter $\rho$}  

Aside from the change from $D^{(k)}$ to \smash{$\tilde{D}^{(x,k)}$},
another key change in the extension of our special ADMM
routine to general inputs $x_1,\ldots x_n$ lies in the choice of
the augmented Lagrangian parameter $\rho$.  Recall that for unit
spacings, we argued for the choice $\rho=\lambda$.  For arbitrary
inputs $x_1 < \ldots < x_n$, we advocate the use of
\begin{equation}
\label{eq:rhox}
\rho = \lambda \left(\frac{x_n-x_1}{n}\right)^k.
\end{equation}
Note that this (essentially) reduces to $\rho=\lambda$ when  
$x_1=1,\ldots x_n=n$.  To motivate the above choice of $\rho$,
consider running two parallel ADMM routines on the same
outputs $y_1,\ldots y_n$, but with different inputs: $1,\ldots n$ in 
one case, and arbitrary but evenly spaced $x_1,\ldots x_n$ 
in the other.  Then, setting $\rho=\lambda$ in the first routine,
we choose $\rho$ in the second routine to try to match the first
round of ADMM updates as best as possible, and this leads to $\rho$ as  
in \eqref{eq:rhox}.  In practice, this input-adjusted choice of $\rho$
makes a important difference in terms of the progress of the algorithm.

\section{ADMM algorithms for trend filtering extensions }  
\label{sec:extend}

One of the real strengths of the ADMM framework for solving
\eqref{eq:tf} is that it can be readily adapted to fit
modifications of the basic trend filtering model.  Here we very
briefly inspect some extensions of trend filtering---some of these 
extensions were suggested by \citet{trendfilter}, some by \citet{l1tf}, 
and some are novel to this manuscript. 
Our intention is not to deliver an exhaustive list of such
extensions (as many more can be conjured), or to study 
their statistical properties, but rather to show that the ADMM
framework is a flexible stage for such creative modeling tasks.

\subsection{Sparse trend filtering}

\begin{figure}[htb]
\centering
\includegraphics[width=0.32\textwidth]{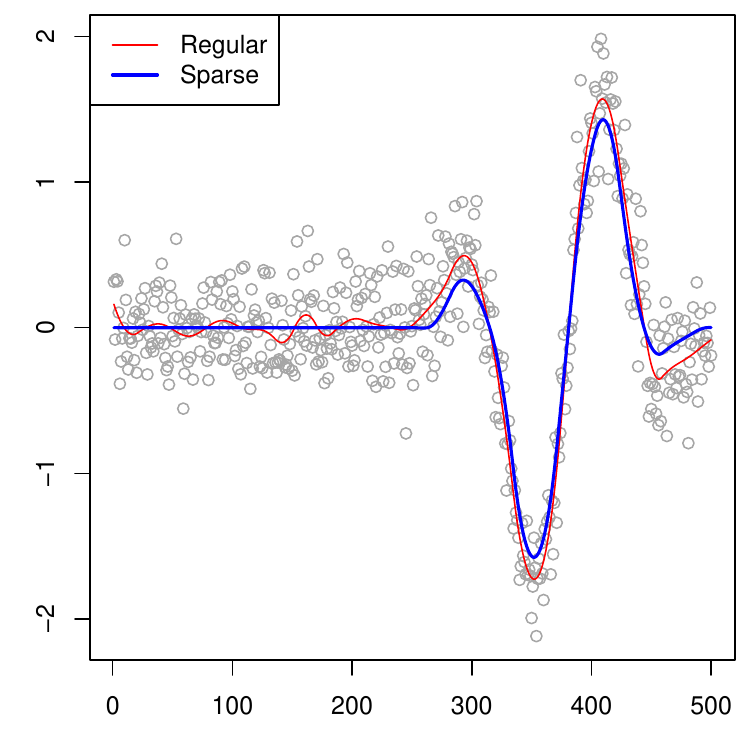} 
\includegraphics[width=0.32\textwidth]{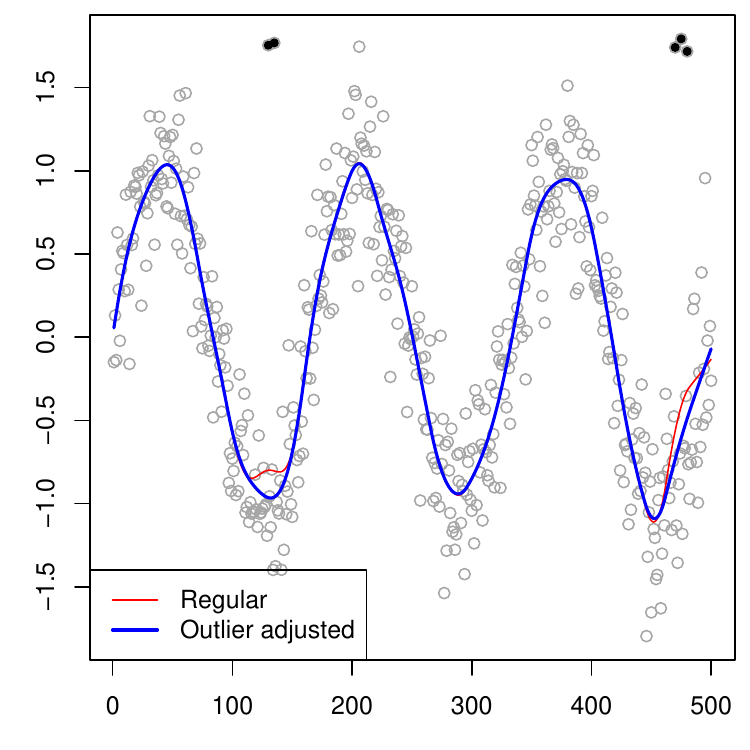} 
\includegraphics[width=0.32\textwidth]{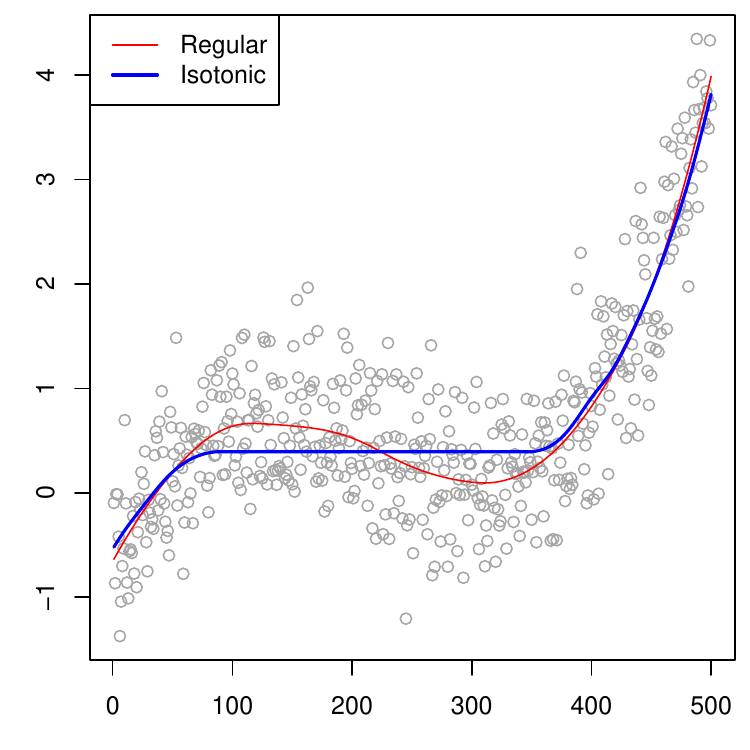} 
\caption{\it\small Three examples, of sparse, outlier-corrected, and 
  isotonic trend filtering, from left to right.  These extensions of
  the basic trend filtering model were computed from $n=500$ data
  points; their fits are drawn in blue, and the original (unmodified)
  trend filtering solutions are drawn in red, both using the same
  hand-chosen tuning parameter values.  (In the middle panel, the 
  points deemed outliers by the nonzero entries of \smash{$\hat{z}$} 
  are colored in black.) 
  These comparisons  are not supposed to be statistically fair, but
  rather, illuminate the qualitative differences imposed by the extra
  penalties or constraints in the extensions.} 
\label{fig:extensions}
\end{figure}

In this sparse variant of trend filtering, we aim to estimate a trend
that can be exactly zero in some regions of its domain, and can 
depart from zero in a smooth (piecewise polynomial) fashion.  This may
be a useful modeling tool when the observations $y_1,\ldots y_n$
represent a difference of signals across common input locations.  We
solve, as suggested by \citet{trendfilter},
\begin{equation*}
\hbeta = \argmin_{\beta\in\R^n} \, \half \|y-\beta\|_2^2 + 
\lambda_1 \|D^{(k+1)} \beta\|_1 + \lambda_2 \|\beta\|_1,
\end{equation*}
where both $\lambda_1,\lambda_2$ are tuning parameters.  A short
calculation yields the specialized ADMM updates:
\begin{align*}
\beta &\leftarrow 
\big((1+\rho_2)I + \rho_1 (D^{(k)})^T D^{(k)} \big)^{-1}
\big(y + \rho_1 (D^{(k)})^T (\alpha + u) +
\rho_2 (\gamma+v)\big), \\ 
\alpha &\leftarrow
\DP_{\lambda_1/\rho_1} (D^{(k)}\beta-u), \\ 
\gamma &\leftarrow S_{\lambda_2/\rho_2} (\beta-v), \\ 
u &\leftarrow u + \alpha - D^{(k)}\beta, \;\;\;
v \leftarrow v + \gamma - \beta.
\end{align*}
This is still highly efficient, using $O(n)$ operations per
iteration.  An example is shown in Figure \ref{fig:extensions}.

\subsection{Mixed trend filtering}

To estimate a trend with two mixed polynomial orders 
$k_1,k_2 \geq 0$, we solve
\begin{equation*}
\hbeta = \argmin_{\beta\in\R^n} \, \half \|y-\beta\|_2^2 +  
\lambda_1 \|D^{(k_1+1)} \beta\|_1 + 
\lambda_2 \|D^{(k_2+1)} \beta\|_1,
\end{equation*}
as discussed in \citet{trendfilter}.
The result is that either polynomial trend, of order $k_1$ or $k_2$,
can act as the dominant trend at any location in the domain. 
More generally, for $r$ mixed polynomial orders, $k_\ell \geq 0$, 
$\ell=1,\ldots r$, we replace the penalty with 
\smash{$\sum_{\ell=1}^r \lambda_\ell \|D^{(k_\ell+1)} \beta\|_1$}. 
The specialized ADMM routine naturally extends to this multi-penalty
problem: 
\begin{align*} 
\beta &\leftarrow 
\Big(I + \sum_{\ell=1}^r \rho_\ell 
(D^{(k_\ell)})^T D^{(k_\ell)}\Big)^{-1} 
\Big(y + \sum_{\ell=1}^r \rho_\ell (D^{(k_\ell)})^T 
(\alpha_\ell + u_\ell )\Big), \\ 
\alpha_\ell &\leftarrow
\DP_{\lambda_\ell/\rho_\ell}(D^{(k_\ell)}\beta-u_\ell), 
\;\;\; \ell=1,\ldots r, \\ 
u_\ell &\leftarrow u_\ell + \alpha_\ell - D^{(k_\ell)}\beta, 
\;\;\; \ell=1,\ldots r.
\end{align*}
Each iteration here uses $O(nr)$ operations (recall $r$ is the
number of mixed trends). 

\subsection{Trend filtering with outlier detection}

To simultaneously estimate a trend and detect outliers, we solve
\begin{equation*}
(\hbeta, \hat{z}) = 
\argmin_{\beta,z\in\R^n} \, \half \|y-\beta-z\|_2^2 +  
\lambda_1 \|D^{(k+1)} \beta\|_1 + 
\lambda_2 \|z\|_1,
\end{equation*}
as in \citet{l1tf}, \citet{owenoutlie}, where the nonzero
components of \smash{$\hat{z}$} correspond to adaptively detected  
outliers.  A short derivation leads to the updates: 
\begin{align*}
\begin{pmatrix} \beta \\ z \end{pmatrix}
&\leftarrow 
\left(\begin{array}{cc}
I + \rho_1 (D^{(k)})^T D^{(k)} & I \\
I & (1+\rho_2)I 
\end{array}\right)^{-1}
\left(\begin{array}{c}
y + \rho_1 (D^{(k)})^T (\alpha+u) \\
y + \rho_2(\gamma + v)
\end{array}\right), \\
\alpha &\leftarrow
\DP_{\lambda_1/\rho_1} (D^{(k)}\beta-u), \\ 
\gamma &\leftarrow S_{\lambda_2/\rho_2} (z-v), \\  
u &\leftarrow u + \alpha - D^{(k)}\beta, \;\;\;
v \leftarrow v + \gamma - z.
\end{align*}
Again, this routine uses $O(n)$ operations per
iteration. See Figure \ref{fig:extensions} for an example.  

\subsection{Isotonic trend filtering}

A monotonicity constraint in the estimated trend is straightforward to
encode: 
\begin{equation*}
\hbeta = \argmin_{\beta\in\R^n} \, \half \|y-\beta\|_2^2 + 
\lambda \|D^{(k+1)} \beta\|_1 \;\;\st\;\; 
\beta_1 \leq \beta_2 \leq \ldots \leq \beta_n,
\end{equation*}
as suggested by \citet{l1tf}.
The specialized ADMM updates are easy to derive: 
\begin{align*}
\beta &\leftarrow 
\big((1+\rho_2)I + \rho_1 (D^{(k)})^T D^{(k)} \big)^{-1}
\big(y + \rho_1 (D^{(k)})^T (\alpha + u) +
\rho_2 (\gamma+v)\big), \\ 
\alpha &\leftarrow
\DP_{\lambda/\rho} (D^{(k)}\beta-u), \\ 
\gamma &\leftarrow \mathrm{IR} (\beta-v), \\ 
u &\leftarrow u + \alpha - D^{(k)}\beta, \;\;\;
v \leftarrow v + \gamma - \beta.
\end{align*}
where $\mathrm{IR}(z)$ denotes an isotonic
regression fit on $z$; since this takes $O(n)$ time (e.g.,
\citet{stout2008}), a round of updates also takes $O(n)$ time. 
Figure \ref{fig:extensions} gives an example.   

\subsection{Nearly-isotonic trend filtering}

Instead of enforcing strict monotonicity in the fitted values, we can
penalize the pointwise nonmontonicities with a separate penalty,
following \citet{neariso}: 
\begin{equation*}
\hbeta = \argmin_{\beta\in\R^n} \, \half \|y-\beta\|_2^2 + 
\lambda_1 \|D^{(k+1)} \beta\|_1 + 
\lambda_2 \sum_{i=1}^{n-1} (\beta_i-\beta_{i+1})_+.
\end{equation*}
This results in a ``nearly-isotonic'' fit \smash{$\hbeta$}.
Above, we use $x_+=\max\{x,0\}$ to denote
the positive part of $x$.  The specialized ADMM updates are:
\begin{align*}
\beta &\leftarrow 
\big((1+\rho_2)I + \rho_1 (D^{(k)})^T D^{(k)} \big)^{-1}
\big(y + \rho_1 (D^{(k)})^T (\alpha + u) +
\rho_2 (\gamma+v)\big), \\ 
\alpha &\leftarrow
\DP_{\lambda_1/\rho_1} (D^{(k)}\beta-u), \\ 
\gamma &\leftarrow \DP_{\lambda_2/\rho_2}^+ (\beta-v), \\  
u &\leftarrow u + \alpha - D^{(k)}\beta, \;\;\;
v \leftarrow v + \gamma - \beta.
\end{align*}
where $\DP_t^+(z)$ denotes a nearly-isotonic regression fit to $z$,
with penalty parameter $t$.  It can be computed in $O(n)$ time by
modifying the dynamic programming algorithm of \citet{nickdp} for the
1d fused lasso, so one round of updates still takes $O(n)$ time.

\section{Conclusion}
\label{sec:discuss}

We proposed a specialized but simple ADMM
approach for trend filtering, leveraging the strength of extremely
fast, exact solvers for the special case $k=0$ (the 1d fused lasso
problem) in order to solve higher order problems with $k \geq 1$.
The algorithm is fast and robust over a wide range of problem sizes
and regimes of regularization parameters (unlike primal-dual interior
point methods, the current state-of-the-art).   
Our specialized ADMM algorithm converges at a far superior rate to
(accelerated) first-order methods, coordinate descent, and (what may
be considered as) the standard ADMM approach for trend filtering.
Finally, a major strength of our proposed algorithm is that it can be
modified to solve many extensions of the basic trend filtering
problem.   
Software for our specialized ADMM algorithm is accessible through the
{\tt trendfilter} function in the R package {\tt glmgen}, built around
a lower level C package, both freely available at  
\url{https://github.com/statsmaths/glmgen}.


\bigskip
\noindent
\textbf{Acknowledgements.} Taylor Arnold and Veeranjaneyulu Sadhanala
contributed tremendously to the development of the new {\tt glmgen}
package.  We thank the anonymous Associate Editor and Referees for
their very helpful reading of our manuscript.
This work was supported by NSF Grant DMS-1309174. 

\bibliographystyle{agsm}  
\bibliography{ryantibs}  

\appendix

\section{Appendix: further details and simulations}

\subsection{Algorithm details for the motivating example}
\label{app:motivate}

\begin{figure}[htbp]
\centering
\includegraphics[width=0.45\textwidth]{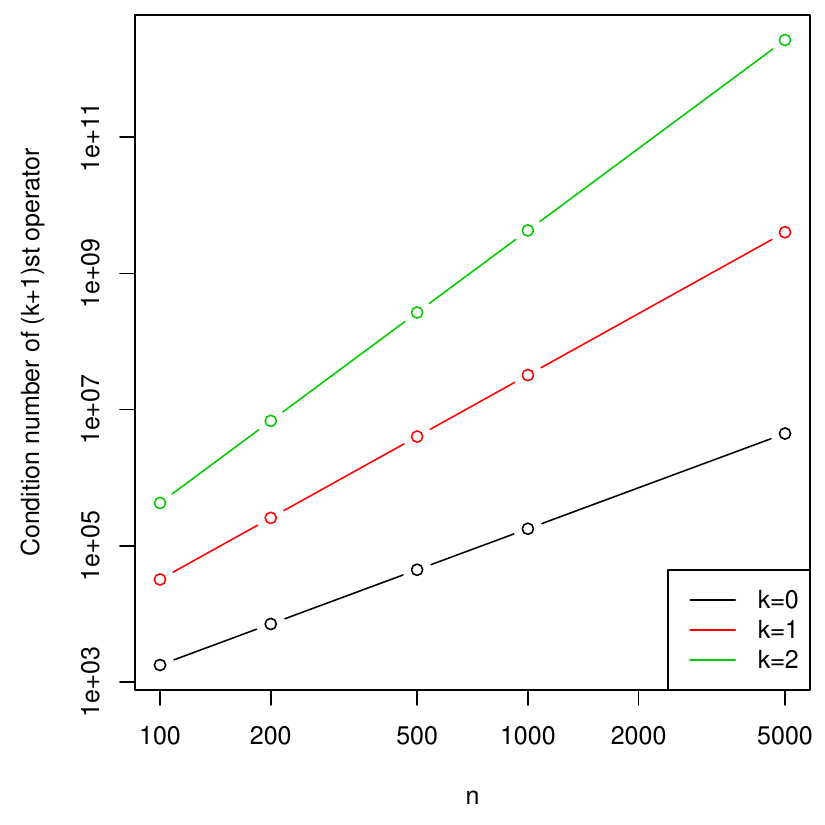}
\caption{\it A log-log plot of the condition number of $D^{(k+1)}$
  versus the problem size $n$, for $k=0,1,2$, where the condition
  numbers scale roughly like $n^k$.}    
\label{fig:cond}
\end{figure}

First, we examine in Figure \ref{fig:cond} the condition numbers of
the discrete difference operators 
$D^{(k+1)} \in \R^{(n-k-1) \times n}$, for varying problem sizes $n$,
and $k=0,1,2$.  Since the plot uses a log-log scale, the straight
lines indicate that the condition numbers grow polynomially with $n$
(with a larger exponent for larger $k$).  The sheer size of the
condition numbers (which can reach $10^{10}$ or larger, even for a
moderate problem size of $n=5000$) is worrisome from an optimization
point of view; roughly speaking, we would expect the criterion in 
these cases to be very flat around its optimum. 

Figure \ref{fig:motivate} (in the introduction) provides evidence that
such a worry can be realized in practice, even with only a reasonable
polynomial order and moderate problem size. 
For this example, we drew $n=1000$ points from an underlying
piecewise linear function, and studied computation of the linear
trend filtering estimate, i.e., with $k=1$, when $\lambda=1000$.   We
chose this tuning parameter value because it represents a
statistically reasonable level of regularization in the example.  The  
{\it exact solution} of the trend filtering problem
at $\lambda=1000$ was computed using the generalized lasso
dual path algorithm \citep{genlasso,fastgenlasso}.  The problem size
here is small enough that this algorithm, which tracks the solution in 
\eqref{eq:tf} as $\lambda$ varies continuously from $\infty$ to 0,
can be run effectively; however, for larger problem sizes, computation
of the full solution path quickly becomes intractable.  Each panel
of Figure \ref{fig:motivate} plots the simulated data points, and
the exact solution as a reference point.  The results of using various
algorithms to solve \eqref{eq:tf} at $\lambda=1000$ are also shown.
Below we give the details of these algorithms.

\begin{itemize}
\item Proximal gradient algorithms cannot be used directly to solve 
the primal problem \eqref{eq:tf} (note that evaluating the proximal 
operator is the same as solving the problem itself).  However,
proximal gradient descent can be applied to the dual of
\eqref{eq:tf}.  Abbreviating $D=D^{(k+1)}$, the dual problem can be
expressed as (e.g., see \citet{genlasso})
\begin{equation}
\label{eq:dual}
\hu =  \argmin_{u\in\R^{n-k-1}} \, \|y-D^T u\|_2^2 \;\; \st
\;\; \|u\|_\infty \leq \lambda.
\end{equation}
The primal and dual
solutions are related by \smash{$\hbeta=y-D^T \hu$}.  We ran
proximal gradient and accelerated proximal gradient descent on
\eqref{eq:dual}, and computed primal solutions accordingly.  Each
iteration here is very efficient and requires $O(n)$ operations, as
computation of the gradient involves one multiplication by $D$ and 
one by $D^T$, which takes linear time since these matrices
are banded, and the proximal operator is
simply coordinate-wise truncation (projection onto an $\ell_\infty$
ball).  The step sizes for each algorithm were hand-selected to be the 
largest values for which the algorithms still converged; this was
intended to give the algorithms the best possible performance.  The
top left panel of 
Figure \ref{fig:motivate} shows the results after
10,000 iterations of proximal gradient its accelerated version
on the dual \eqref{eq:dual}.  The fitted curves are
wiggly and not piecewise linear, even after such an unreasonably large
number of iterations, and even with acceleration (though
acceleration clearly provides an improvement). 

\item The trend filtering problem in \eqref{eq:tf} can alternatively
  be written in lasso form,
\begin{equation}
\label{eq:lasso}
\htheta = \argmin_{\theta \in \R^n} \,
\half \|y - H \theta\|_2^2 + \lambda \cdot k! 
\sum_{j=k+2}^n |\theta_j|, 
\end{equation}
where $H = H^{(k)} \in \R^{n \times n}$ is $k$th order 
falling factorial basis matrix, defined over $x_1,\ldots x_n$, which,
recall, we assume are $1,\ldots n$.  The matrix $H$ is effectively the 
inverse of $D$ \citep{trendfilter}, and the solutions of \eqref{eq:tf}  
and \eqref{eq:lasso} obey \smash{$\hbeta=H\htheta$}.  The lasso
problem \eqref{eq:lasso} provides us with another avenue for proximal
gradient descent.  Indeed the iterations of proximal gradient descent
on \eqref{eq:lasso} are very efficient and
can still be done in $O(n)$ time: the gradient computation requires 
one multiplication by $H$ and $H^T$, which can be applied in linear
time, despite the fact that these matrices are dense \citep{fallfact},
and the proximal map is coordinate-wise soft-thresholding.  
After 10,000 iterations, as we can see from the top right panel of
Figure \ref{fig:motivate}, this method still gives an unsatisfactory fit,
and the same is true for 10,000 iterations with acceleration
(the output here is close, but it is not piecewise linear, having
rounded corners).

\item The bottom left panel in the figure explores two commonly used
  non-first-order methods, namely, coordinate descent applied to the
  lasso formulation \eqref{eq:lasso}, and a standard ADMM approach on
  the original formulation \eqref{eq:tf}.  The standard ADMM algorithm
  is described in Section \ref{sec:admm}, and has $O(n)$ per iteration 
  complexity.  As far as we can tell, coordinate descent requires
  $O(n^2)$ operations per iteration (one iteration being a full
  cycle of coordinate-wise minimizations), because the update rules
  involve multiplication by individual columns of $H$, and not $H$
  in its entirety.  The plot shows the results of these two algorithms
  after 5000 iterations each.  After such a large number of
  iterations, the standard ADMM result is fairly close to the
  exact solution in some parts of the domain, but overall fails to
  capture the piecewise linear structure.  Coordinate descent, on the
  other hand, is quite far off (although we note that it does deliver
  a visually perfect piecewise linear fit after nearly 100,000 iterations).

\item The bottom right panel in the figure justifies the perusal of
  this paper, and should generate excitement in the
  curious reader. It illustrates that after just {\it 20 iterations},
  both the PDIP method of \citet{l1tf}, and our special
  ADMM implementation deliver results that are visually
  indistinguishable 
  from the exact solution.  In fact, after only $5$ iterations, 
  the specialized ADMM fit (not shown) is visually passable.    
  Both algorithms use $O(n)$ operations per iteration:
  the PDIP algorithm is actually applied to the dual problem
  \eqref{eq:dual}, and its iterations reduce to solving linear systems
  in the banded matrix $D$; the special ADMM algorithm in described
  in Section \ref{sec:admm}.
\end{itemize}

\subsection{ADMM vs. PDIP for $k=3$   
(piecewise cubic fitting)}  
\label{app:k3}

\begin{figure}[h!]
\centering
\includegraphics[width=0.32\textwidth]{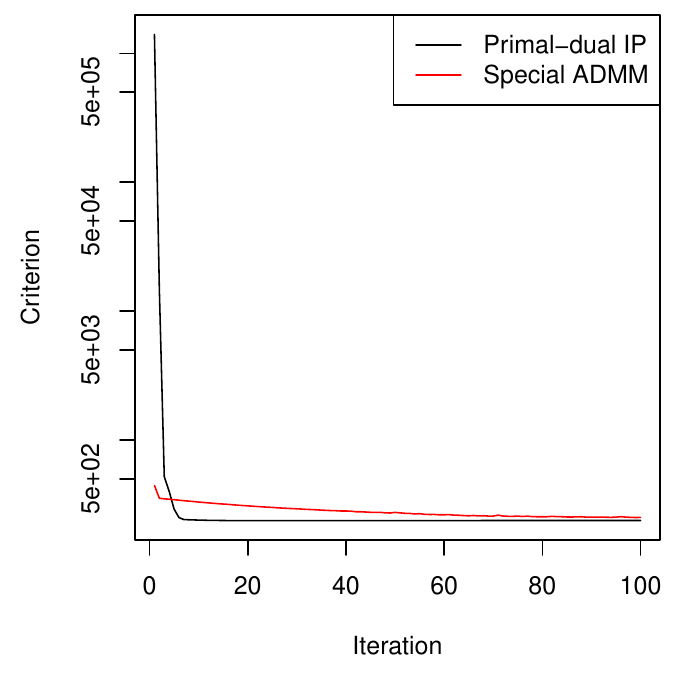} 
\includegraphics[width=0.32\textwidth]{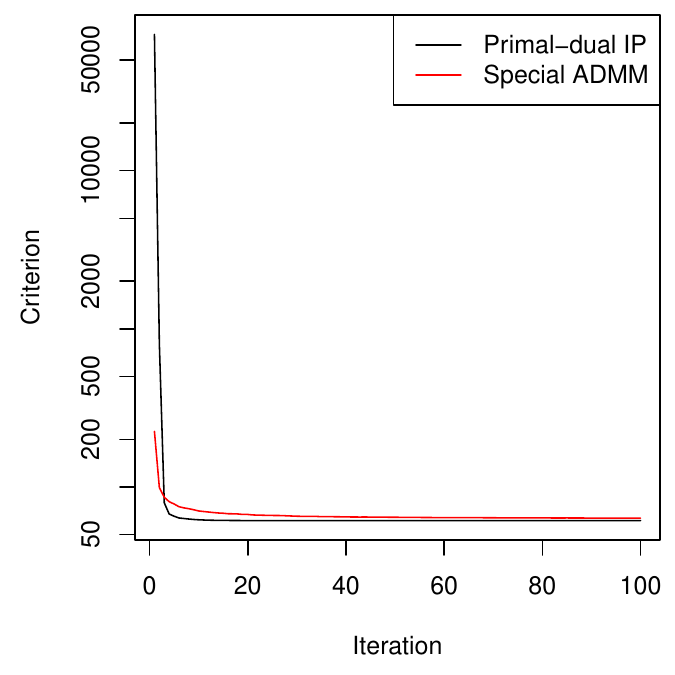}
\includegraphics[width=0.32\textwidth]{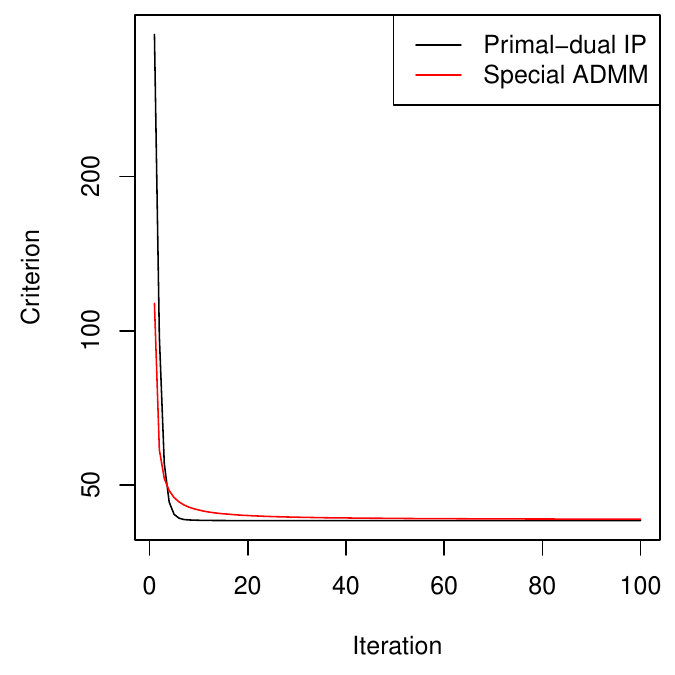}\\ 
\includegraphics[width=0.32\textwidth]{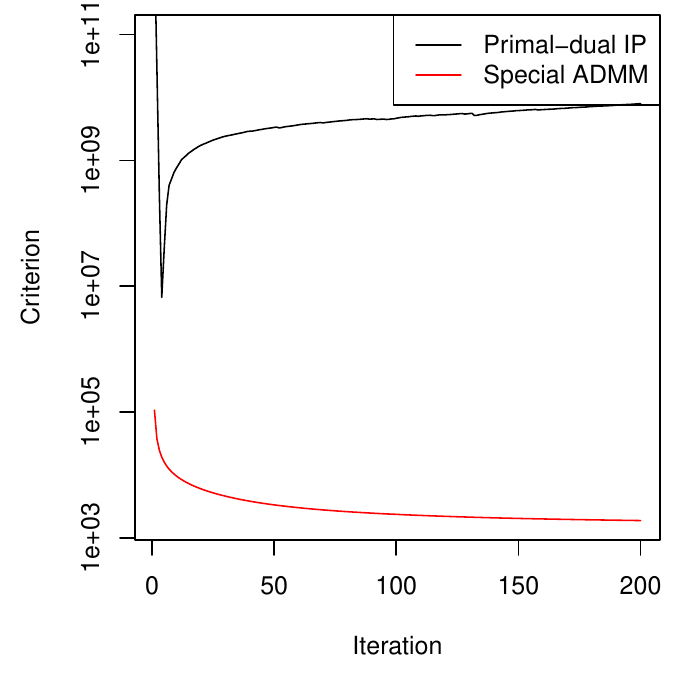} 
\includegraphics[width=0.32\textwidth]{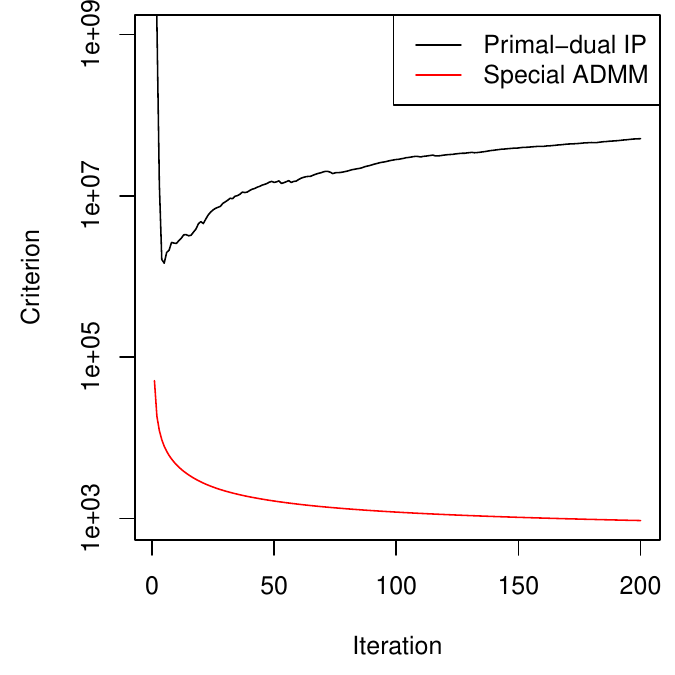}
\includegraphics[width=0.32\textwidth]{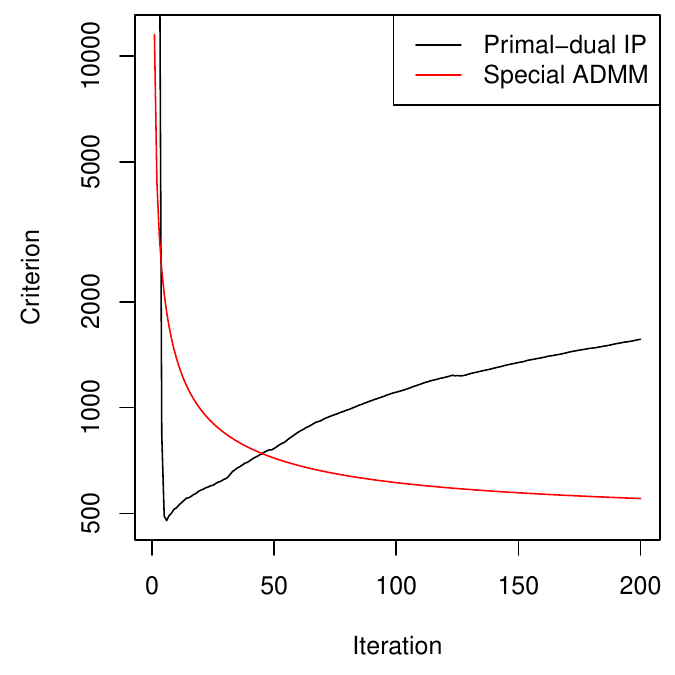} 
\caption{\small\it Convergence plots for ($k=3$): achieved criterion 
  values across iterations of ADMM and PDIP, with the same layout as
  in Figures \ref{fig:crit-pd-admm-k1-conv} and
  \ref{fig:crit-pd-admm-k2-conv}, except that the first row
  uses $n=1000$ points, and the second row $n=10,000$ points. 
  Both algorithms comfortably converge when $n=1000$.
  However, PDIP encounters serious difficulties  
  when $n=10,000$, reminiscent of its behavior for $k=2$ but when
  $n=100,000$
  (see Figure \ref{fig:crit-pd-admm-k2-conv}).  
  In all cases, the specialized ADMM algorithm
  demonstrates a strong and steady convergence behavior.
  }
\label{fig:crit-pd-admm-k3}
\end{figure}

For the case $k=3$ (piecewise cubic fitting), the behavior of PDIP
mirrors that in the $k=2$ case, yet the convergence issues begin to
show at problem sizes smaller by an order of magnitude.  
The specialized ADMM approach is slightly slower to converge, but
overall still quite fast and robust.  Figure \ref{fig:crit-pd-admm-k3}
supports this point.  

\subsection{Prediction at arbitrary points}
\label{app:predict}

Continuing within the nonparametric regression context, 
an important task to consider is that of function prediction 
at arbitrary locations in the domain. We discuss how to make such
predictions using trend filtering.  This
topic is not directly relevant to our particular algorithmic proposal,
but our R software package that implements this algorithm 
also features the function prediction task, and hence we describe it
here for  completeness.  The trend filtering
estimate, as defined in \eqref{eq:tf}, produces fitted values 
\smash{$\hbeta_1,\ldots \hbeta_n$} at the given input points 
$x_1,\ldots x_n$.  We may think of these fitted values as the
evaluations of an underlying fitted function \smash{$\hf$}, as in      
$\big(\hf(x_1), \ldots \hf(x_n)\big) = 
(\hbeta_1,\ldots \hbeta_n). $
\citet{trendfilter}, \citet{fallfact} argue that the appropriate 
extension of \smash{$\hf$} to the continuous domain is given by 
\begin{equation}
\label{eq:hf}
\hf(x) = \sum_{j=1}^{k+1} \hat{\phi}_j \cdot h_j(x) + 
\sum_{j=1}^{n-k-1} \htheta_j \cdot h_{k+1+j}(x),
\end{equation}
where $h_1,\ldots h_n$ are the falling factorial basis functions,
defined as
\begin{gather*}
h_j(x) = \prod_{\ell=1}^{j-1} (x-x_\ell), \;\;\; j=1,\ldots k+1, \\
h_{k+1+j}(x) = \prod_{\ell=1}^k (x-x_{j+\ell}) \cdot 
1 \{x \geq x_{j+k}\}, \;\;\; j=1,\ldots n-k-1,
\end{gather*}
and \smash{$\hat{\phi} \in \R^{k+1}$}, 
\smash{$\htheta \in \R^{n-k-1}$} are inverse coefficients to
\smash{$\hbeta$}.  The first $k+1$ coefficients index the polynomial
functions $h_1,\ldots h_{k+1}$, and defined by 
\smash{$\hat{\phi}_1 = \hbeta_1$}, and
\begin{equation}
\label{eq:phi}
\hat{\phi}_j = \frac{1}{(j-1)!} \cdot \left[
\diag\left( \frac{1}{x_j-x_1}, \ldots \frac{1}{x_n-x_{n-j+1}} \right)
\cdot D^{(x,j-1)} \right]_1 \cdot \hbeta,
\;\;\; j=2,\ldots k+1.
\end{equation}
Above, we use $A_1$ to denote the first row of a matrix $A$.  Note 
that \smash{$\hat{\phi}_1,\ldots \hat{\phi}_{k+1}$} are generally
nonzero at the trend filtering solution \smash{$\hbeta$}.  The last
$n-k-1$ coefficients index the knot-producing functions
$h_{k+2},\ldots h_n$, and are defined by
\begin{equation}
\label{eq:theta}
\htheta = D^{(x,k+1)} \hbeta / k!.
\end{equation}
Unlike \smash{$\hat\phi$}, it is apparent that many of 
\smash{$\htheta_1,\ldots \htheta_{n-k-1}$} will be zero at the trend
filtering solution, more so for large $\lambda$. Given a trend
filtering estimate \smash{$\hbeta$}, we can precompute the
coefficients \smash{$\hat{\phi},\htheta$} as in \eqref{eq:phi}.  Then,
to produce evaluations of the underlying
estimated function \smash{$\hf$} at arbitrary points 
\smash{$x'_1,\ldots x'_m$}, we calculate the linear combinations
of falling factorial basis functions according to \eqref{eq:hf}.
From the precomputed coefficients \smash{$\hat{\phi},\htheta$}, this
requires only $O(mr)$ operations, where
\smash{$r=\|D^{(x,k+1)}\hbeta\|_0$}, the number of nonzero $(k+1)$st
order differences at the solution (we are taking $k$ to be a
constant).

\end{document}